\documentclass[11pt]{article}

\usepackage[]{EMNLP2023}

\usepackage{times}
\usepackage{latexsym}
\usepackage{makecell}
\usepackage[T1]{fontenc}

\usepackage[utf8]{inputenc}

\usepackage{microtype}

\usepackage{inconsolata}
\usepackage{graphicx}
\usepackage{soul}
\usepackage{multicol}
\usepackage{multirow}
\usepackage{amssymb}

\usepackage{microtype}

\usepackage{multirow}
\usepackage{xcolor,colortbl}
\usepackage{amsmath,array,graphicx}
\usepackage{hhline}
\usepackage{rotating}
\usepackage{graphicx}

\title{Zero-shot Cross-lingual Transfer Learning with Multiple Source and Target Languages for Information Extraction: Language Selection and Adversarial Training}

\author{Nghia Trung Ngo \and Thien Huu Nguyen \\
  Department of Computer Science \\
  University of Oregon, Eugene, Oregon, USA \\
  \texttt{nghian@uoregon.edu, thien@cs.uoregon.edu} 
  \\}

\begin{document}
\maketitle
\begin{abstract}
The majority of previous researches addressing multi-lingual IE are limited to zero-shot cross-lingual single-transfer (one-to-one) setting, with high-resource languages predominantly as source training data. 
As a result, these works provide little understanding and benefit for the realistic goal of developing a multi-lingual IE system that can generalize to as many languages as possible.
Our study aims to fill this gap by providing a detailed analysis on Cross-Lingual Multi-Transferability (many-to-many transfer learning), for the recent IE corpora that cover a diverse set of languages.
Specifically, we first determine the correlation between single-transfer performance and a wide range of linguistic-based distances.
From the obtained insights, a combined language distance metric can be developed that is not only highly correlated but also robust across different tasks and model scales.
Next, we investigate the more general zero-shot multi-lingual transfer settings where multiple languages are involved in the training and evaluation processes. 
Language clustering based on the newly defined distance can provide directions for achieving the optimal cost-performance trade-off in data (languages) selection problem.
Finally, a relational-transfer setting is proposed to further incorporate multi-lingual unlabeled data based on adversarial training using the relation induced from the above linguistic distance.
Experimental results on two practical multi-lingual IE tasks demonstrate our method significantly outperforms baselines across tasks and languages simultaneously.
Additionally, by carefully designing the multi-lingual training to utilize data from relevant languages, we can achieve a substantial boost in generalization ability with reasonable labor cost for the additional data collection.
\end{abstract}

\section{Introduction}

The objective of Information extraction (IE) is to identify and extract structure information, such as entities, relations, and events, from natural unstructured text. IE plays an important role in various downstream applications, including Question Answering, Knowledge Graph Construction, News Analysis, etc. 
Solving IE tasks pose significant challenges for NLP models as they often require the understanding of complex features of natural languages.
For example, to extract relations within a sentence, models first need to learn specialized structures of the corresponding language to identify entities mentioned in the given text. Next, a deep understanding of context is required to correctly classify the relations between these entities.
These challenges are further exacerbated in multilingual settings, where datasets are collected from multiple languages, each of which contains language-specific characteristics and structures.

The rapid development around English-based datasets has pushed machine performance to be on par with human ability in English tasks, prompting recent works to explore NLP research in other languages \cite{xglue, xtr-21}.
However, despite advanced large-scale architectures and high English results, current models notably under-perform in new languages, especially those that are considered low-resource and lack high-quality datasets for ﬁne-tuning.
Cross-lingual Transfer, as a result, becomes one of the most important directions in the field.
Given a particular task, the goal of Cross-lingual Transfer is to train multilingual models over high-resource source languages that can solve textual tasks in new target languages despite the shifts in linguistic origin.

Currently, the most popular and practical approach for IE involves Zero-Shot Cross-Lingual (ZSCL) transfer \cite{xlmr, xlmr-xl}. These methods fine-tune Transformer-based  multilingual Language Models (mLMs), which were pre-trained using unlabeled text from hundreds of languages, for downstream tasks using high-resource source-language labeled datasets (predominantly English). The resulting models are directly used for evaluation on the corresponding tasks in target languages.
Studies have shown, however, performance of these multi-lingual models varies substantially across languages and tasks.
Several factors have been attributed to this phenomenon, ranging from data-dependent statistic(e.g. dataset size, word overlap) \cite{malkin:22}, to data-independent features (e.g. phylogenetic and typological features) \cite{lin:19, dolicki:21}.
Based on these previous observations, we believe that there is a deeper connection between cross-lingual transfer ability and the relations in the linguistic landscape. Unraveling this correlation can provide tremendous practical implications for IE.
First, it serves as a guide for data collection process to achieve optimal cost-performance trade-off by gathering training samples from appropriate source languages for a target language. Furthermore, the modeling process can also be tailored such that the learned representations explicitly capture the linguistic relations to improve generalization across languages.

Previous papers following the above direction define the problem as a Performance Prediction task. In \cite{lin:19, dolicki:21, srinivasan:21}, a regression model is trained to take linguistic features of the source-target language pairs as input to predict a trained model performance scores on target languages.
Despite high prediction accuracy, these works are insufficient for the following two reasons.
First, they place too much emphasis on the accuracy of the regression model, which is trained for a specific architecture on a particular task.
As the training configuration varies widely in practice, the results obtained from the performance prediction models may become unreliable and not applicable in general.
Another reason is that previous work is only limited to the setting of single-transfer between two languages, in which only one source language (predominately English) is utilized.
Current advances in translation model and data gathering process have enabled the creation of datasets in many languages, thus multiple source languages should be considered.
Intuitively, additional training data from more languages can help improve model's generalization on downstream tasks, and learning from text in multiple languages may have a positive effect on zero-shot transfer.
We believe that multi-transfer setting is the next important step for cross-lingual transfer, both to improve model performance across languages and to provide a more complete picture of multi-linguality in machine learning.
In this paper, we focus on what has been missing in previous works by aiming to answer the following three major research questions:

\textbf{Q1: How do the relations in the linguistic landscape affect an IE model’s cross-lingual transfer ability?}  
We use URIEL Typological Database \cite{uriel} to extract phylogenetic and typological properties of each language.
These properties, represented as multi-dimensional binary features, are used to compute the pair-wise linguistic distances (or equivalently the similarity scores) between languages.
We compare the correlation between these scores and model single-transfer performance.
A source language with a high correlation value would imply that we can infer its ability to transfer to different languages using only linguistic relations, without the need to actually fine-tune models.

\textbf{Q2: Can we implicitly leverage these linguistic features as dataset-independent knowledge to efficiently address the more general multi-transfer setting?}
While many-to-many cross-lingual transfer has the potential to significantly improve single-transfer performance, it would also require gathering data from multiple languages.
Given a set of languages and their linguistic features as the only prior information, we aim to find the optimal subset of source languages to gather labeled data for zero-shot cross-lingual multi-transfer to target languages.
The goal is to achieve the best cost-performance trade-off on all languages, without having to fine-tune the mLMs on an exponential number of possible language combinations.

\textbf{Q3: Can we explicitly integrate these linguistic relations in the learning process to effectively improve multi-transfer performance?} 
We then investigate further into the possibility of directly embedding the linguistic relations in the fine-tuning process.
The hypothesis is that, by capturing these connections, the multi-lingual representations would be able to adaptatively generalize to not only languages that are closely related to source languages, but also distant languages that share little similarity with the available training data.

The following observations are obtained from our quantitative experiments and qualitative analysis, through 3 levels of transfer settings:

\textbf{1) Single-transfer (ZSCL-S) - Only 1 labeled source language available.}
It is possible to achieve a high degree of correlation between model ZSCL performances and linguistic relations, using a combination of syntax, inventory, and phonology features from URIEL.
However, in contrast to prior works which only focus on syntactic transfer when fine-tuning, our combined metric places the least importance on the syntax feature. 
This implies that previous researches are suboptimal and incomplete, prompting further investigations into the problem.

\textbf{2) Multi-transfer (ZSCL-M) - Multiple labeled source language available.}
We first cluster languages based on the combined metric above. Then, by selecting source languages following the guidance from the resulting clusters, we observe significant improvements in ZSML performances over the naive method of randomly picking source languages.
In other words, with only the prior linguistic knowledge, we can efficiently choose a suitable small subset of languages for labeled data annotations, to fine-tune a MMLM to perform best on a given set of languages.

\textbf{3) Relational-transfer (ZSCL-R) - ZSCL-M with additional multi-lingual unlabeled data.}
We leverage unlabeled data from all available languages and their linguistic relations as inputs to graph-relational adversarial learning framework \cite{grda}, a generalization of adversarial language adaption \cite{sentiment} that can only perform strict uniform alignment for pair-wise transfer.
By conditioning the multi-lingual representation flexibly on the connections expressed by the corresponding language relational-graph, we achieve a considerable increase in transfer performances across every language. This is only at the small cost of collecting additional unlabeled data from other languages.

\begin{table}[]
  \centering
  \small
\captionsetup{skip=2pt}
\addtolength{\belowcaptionskip}{-4mm}
\resizebox{0.43\columnwidth}{!}{
\begin{tabular}{lr|lr}
SMILER                      & \%                            & MINION                         & \%                            \\ \hline
ita                         & 19.71                         & \cellcolor[HTML]{F4CCCC}eng & \cellcolor[HTML]{F4CCCC}39.76 \\
fra                         & 16.25                         & \cellcolor[HTML]{F4CCCC}pol & \cellcolor[HTML]{F4CCCC}13.7  \\
deu                         & 13.75                         & tur                         & 13.7                          \\
\cellcolor[HTML]{F4CCCC}por & \cellcolor[HTML]{F4CCCC}11.54 &                             & \multicolumn{1}{l}{}          \\
nld                         & 10.38                         &                             & \multicolumn{1}{l}{}          \\ \hline
\rowcolor[HTML]{F4CCCC} 
eng                         & 9.57                          & \cellcolor[HTML]{D9EAD3}spa & \cellcolor[HTML]{D9EAD3}9.99  \\
\rowcolor[HTML]{F4CCCC} 
kor                         & 5                             & por                         & 4.59                          \\
\rowcolor[HTML]{F4CCCC} 
pol                         & 4.5                           & swe                         & 4.59                          \\
\cellcolor[HTML]{D9EAD3}spa & \cellcolor[HTML]{D9EAD3}2.95  &                             & \multicolumn{1}{l}{}          \\
ara                         & 2.49                          &                             & \multicolumn{1}{l}{}          \\ \hline
rus                         & 1.71                          & hin                         & 4.58                          \\
\rowcolor[HTML]{F4CCCC} 
swe                         & 1.2                           & kor                         & 4.58                          \\
fas                         & 0.7                           & jpn                         & 4.5                           \\
ukr                         & 0.26                          &                             & \multicolumn{1}{l}{}         
\end{tabular}
}
\caption{\small{Percentage distributions of training data in each task for every language, which are separated into high, medium, and low resource categories. The shared languages are color-coded, with red indicating that the language belongs to a different category between the two tasks, whereas green indicates otherwise. This study involves a total of 17 languages including: arabic (ara), german (deu), english (eng), farsia (fas), french (fra), hindi (hin), italian (ita), japanese (jpn), korean (kor), dutch (nld), polish (pol), portuguese (por), russian (rus), spanish (spa), swedish (swe), turkish (tur), and ukrainian (ukr).}}
\label{tab:datasize}
\end{table}

\begin{figure}[ht!]
\captionsetup{skip=2pt}
\addtolength{\belowcaptionskip}{-7mm}
\begin{center} 
\includegraphics[width=0.36\textwidth]{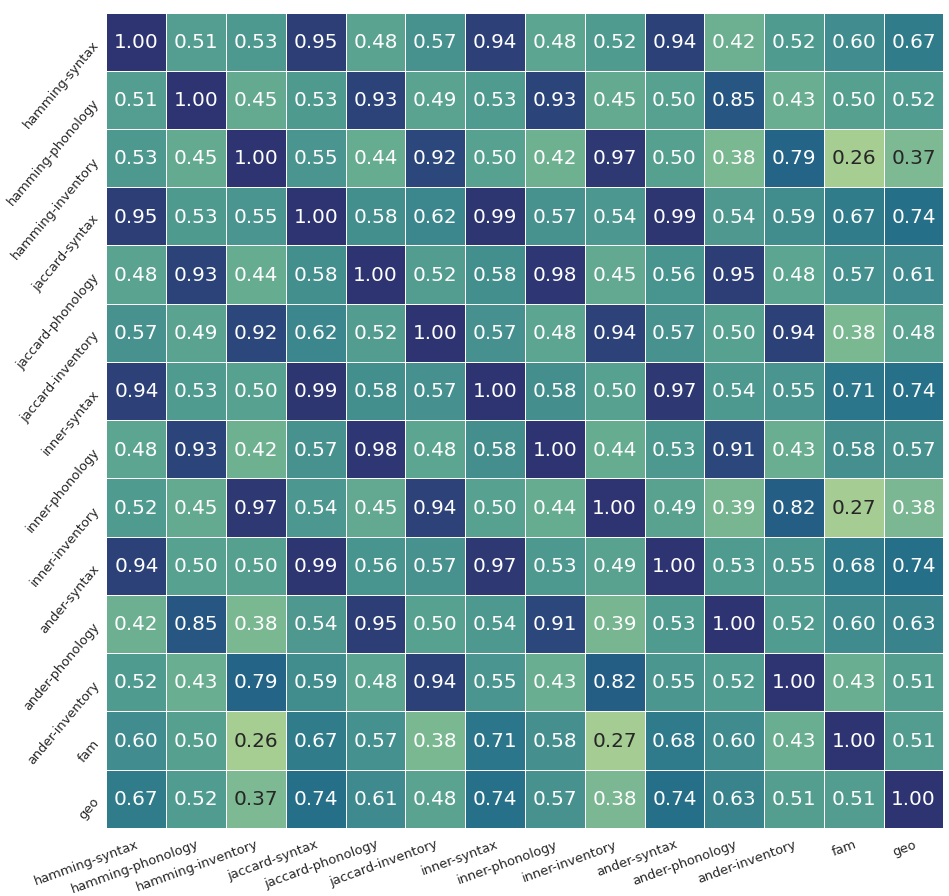}
\caption{\small The pairwise Pearson correlation for all computed language distances.} \label{fig:DistCorr}
\end{center}
\end{figure}

\section{Related Work}

\textbf{Zero-shot Cross-lingual Transfer}
The majority of recent ZSCL works \cite{filter, infoxlm} follow single-transfer setting, using comprehensive multi-lingual multi-task benchmarks such as XTREME \cite{xtr-21}, or XGLUE \cite{xglue}.
These datasets provide English training data for fine-tuning the pre-trained MMLMs, which are then evaluated on translated test sets in different languages.
As a result, English becomes the dominant source language for transfer in following ZSCL researches \cite{phang20}, which is suboptimal due to the linguistic diversity of languages.
Specifically, \cite{keung20} discovers that model's Engish dev accuracy does not correlate with its performance of other languages, and \cite{hero} demonstrates the limitation of using English to transfer to low-resource languages. Furthermore, \cite{primacy} find that other high-resource languages such as German and Russian often transfer more effectively when the set of target languages is diverse or unknown a priori.
Our work builds on these findings and provides a more complete view of language relations and ZSCL performances of mLMs.

\textbf{Linguistic Diversity}
By probing the learned representation of mLMs,  \cite{pires19, limisiewicz20} have found syntactical information implicitly encoded in layers of the multi-lingual models (typically at middle-level layers of the architectures). 
\cite{ningyu22} shows that the pre-training and fine-tuning processes transform these features and directly impact model multi-lingual performances.
Further investigation by \cite{lin:19} demonstrates that distances based on linguistic features, including phylogenetic and typological properties, between two languages, are correlated with cross-lingual transfer capacity. 
These features can be used to further improve transfer guide parameters sharing among languages \cite{ammar16}, or data selection \cite{ponti18}.
Several works \cite{lin:19, litmus} specifically aim to predict cross-lingual transfer performance directly, without training, only from linguistic distances of languages.
Our work follows their line of reasoning but aims to address their limitation to restricted experiment settings (one-to-one transfer, model architectures, tasks, etc.).
In particular, we focus on building a comprehensive picture of linguistic relations and transfer performances, in the general zero-shot mult-transfer (many-to-many) setting, for practical information extraction tasks.

\textbf{Adversarial Language Learning}
Inspired by domain adversarial neural network (DANN) \cite{dann} from domain adaptation research, Adversarial Language Adaptation (ALA) network can be used to extract language-invariant features useful for downstream tasks across languages.
Several works have successfully adopted ALA for cross-lingual transfer setting for different tasks such as sentiment analysis \cite{sentiment}, information extraction \cite{nguyen21b, ngo21}, and name tagging \cite{nametag}.
We generalize these works to cross-lingual transfer with multiple source-target languages.
This is achieved through graph-relational adversarial learning framework following \cite{grda}, a generalization of DANN.

\begin{figure}[ht!]
\captionsetup{skip=2pt}
\addtolength{\belowcaptionskip}{-7mm}
\begin{center} 
\includegraphics[width=0.42\textwidth]{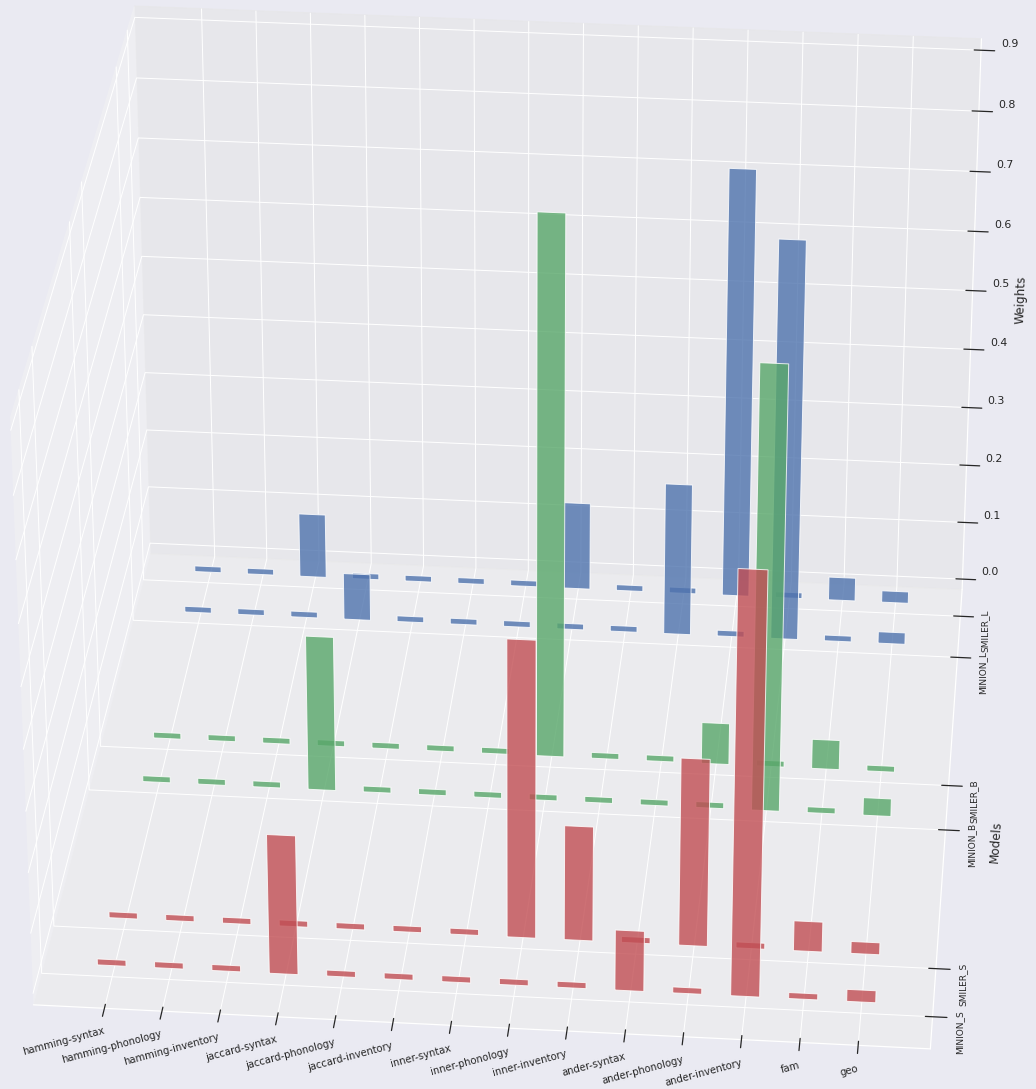}
\caption{\small Feature importance weights of the optimal combined metric for each (task, model scale) setting. Small, base, and large models are represented by the colors red, green, and blue, respectively.} 
\label{fig:LingFeat}
\end{center}
\end{figure}

\section{Datasets}

Information extraction tasks extract structured contextual information from unstructured text, thus requiring model’s comprehension of both syntactic and semantic knowledge in multilingual documents.
In this paper, to demonstrate to the heterogeneity of ZSCL for multi-lingual IE problems in practice, we experiment on two recent datasets that provide training and evaluation data in a wide range of languages.

\textbf{MINION:}
Multi-lingual Event Detection (MINION) \cite{minion:22} annotates event triggers for 8 typologically different languages.
The goal event detection task is to identify the word(s) that describe the occurrence of an event the best from a given text, also referred to as he event trigger, and classify that event into one of the 16 predetermined event types.

\textbf{SMiLER:}
Samsung MultiLingual Entity and Relation Extraction (SMiLER) \cite{smiler:21} consists of annotated entities and relations from 14 languages.
Given an input text, SMiLER not only requires models to identify two entity mentions in the text but also predict their relation from a set of 36 predefined relations.


The distribution of every language training set in each dataset is presented in Table \ref{tab:datasize}.
After categorizing the languages into high/medium/low resources groups, we notice a considerable discrepancy in the categories of shared languages between the two tasks
This reflects the diversity of actual multi-lingual data annotation processes for practical tasks.
Thus, instead of balancing data across languages as in prior studies \cite{malkin:22}, we decide to utilize the original splits of each dataset to demonstrate a realistic picture of cross-lingual transfer performance.

\begin{figure}[ht!]
\captionsetup{skip=2pt}
\addtolength{\belowcaptionskip}{-7mm}
\begin{center} 
\includegraphics[width=0.42\textwidth]{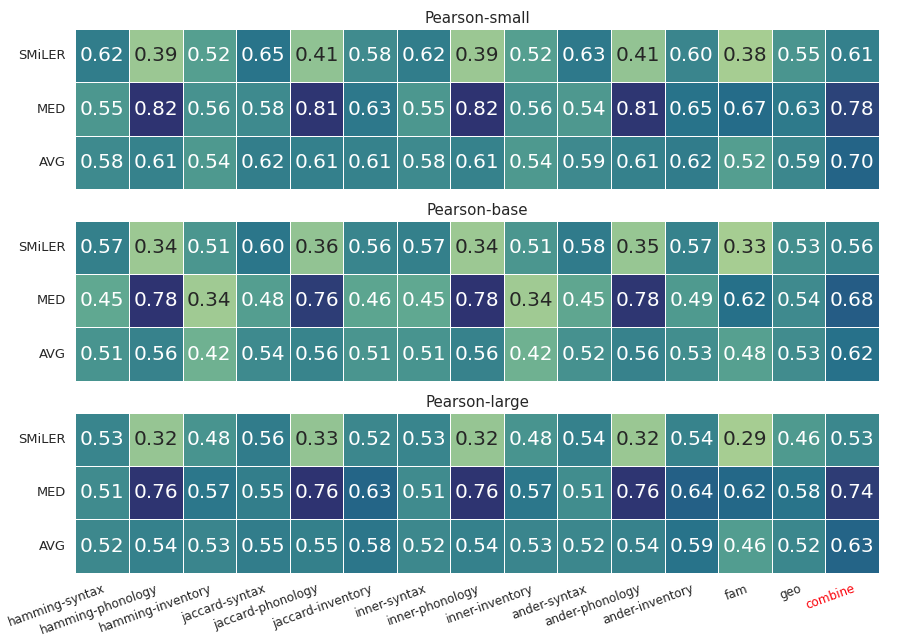}
\caption{\small The language-based average Pearson correlation scores of all computed linguistic distances (including the combined metric).} \label{fig:CombCorr}
\end{center}
\end{figure}

\section{Linguistic Relations} \label{sec:LingRel}
To illustrate a comprehensive picture of the linguistic relations among the available languages, we consider different base linguistic features and how to compare them, into a total of 14 distance metrics

\subsection{Linguistic Features}
Following the standard approach, We use five different linguistic features provided by the URIEL Typological Database \cite{uriel}, including a phylogeny feature, a geography feature, and three typological features (syntax, phonology, and inventory):

\textbf{Phylogeny (fam):} the membership in language families derived from the world language family tree in Glottolog \cite{glottolog}

\textbf{Geography (geo):} the language location based on Glottolog, more specifically the orthodromic distance between the language and a fixed point on the surface of the earth.

\textbf{Syntax:} the language syntactic structures derive from either WALS \cite{wals} or Ethnologue \cite{ethnologue}

\textbf{Phonology:} the phonology features extracted in a similar manner from WALS and Ethnologue

\textbf{Inventory:} the phonetic features derived from PHOIBLE’s phonetic inventories \cite{phoible}

Each of the above linguistic features is represented by a multi-dimensional binary vector for every language, where a value 0 (1) in each dimension represents the absence (presence) of a particular linguistic phenomenon for that language.






\subsection{Distance Metrics}
To calculate the linguistic distance between languages based on the above feature vectors, previous works only consider cosine or Euclidean distance between the binary vectors.
However, numerous binary similarity measures have been proposed and play a critical role in many problems in various fields.
These binary metrics are distinguished by their unique synthetic properties (negative matches, count differences, correlation, etc.), and applying an appropriate one is the key to more accurate data analysis results.
Based on the categorization in \cite{binary} which survey over 76 binary similarity measures, we decide to focus on the following 4 representative distances: Hamming, Jaccard, Inner-product, and Anderberg.

Figure \ref{fig:DistCorr} show the correlations between every pair of our considered linguistic distances (the full detailed distance values are provided in figure \ref{fig:app_lingdist}) in appendix \ref{app:detail}.
Aside from fam and geo which are computed using Euclidean distance, each of the three typological features is computed using the chosen 4 binary metrics, resulting in a total of 14 linguistic distance metrics in the figure.
We can observe significant variations in metrics based on different types of features from the correlation heatmap. In addition, there are also noticeable distinctions between several metrics within the same feature types, particularly between Anderberg and Hamming-based distances.
Noted that in the context of realistic cross-lingual transfer, these two distances maybe provide more insight as transfer performance is asymmetric (Hamming) and non-zero self-distance (Anderberg).
Overall, our choice of linguistic distances ensures a high diversity of correlation that can be computed from the language features.
\section{Zero-shot Cross-lingual Single-transfer}


To answer the first research question, we evaluate ZSCL-S scores for every language pair of each task, in three model scales: small (MiniLM \cite{minilm}), base (XLM-Roberta-base \cite{xlmr}), and large (XLM-Roberta-large).
Next, Pearson correlations are computed between the transfer performances and linguistic distances to identify the degree to which  the relations in the linguistic landscape determine a model’s cross-lingual transfer ability.

\textbf{Experimental Setup} In ZSCL-S, given a pair of source and target languages, the model is trained using labeled data from source language. The ZSCL-S score is then defined as the zero-shot evaluation of the trained model on the test set of target language.

\textbf{Transfer Performance}
Detailed transfer scores are provided in figures \ref{fig:app_zscls_m} and \ref{fig:app_zscls_s} in appendix \ref{app:detail}.
While the language-wise order of the transfer scores is maintained across different model sizes, it is not clear, however, if language identities alone are able to determine model cross-lingual transfer ability.
This is due to the significant difference between the results of the two tasks.
Even more unexpectedly, model transfer scores do not increase linearly with its number of parameters

\subsection{Linguistic Correlation}
We determined if any of the linguistic distances defined in section \ref{sec:LingRel} can explain the heterogeneity of resulting transfer performances, across all settings.

\textbf{Distance-Transfer Correlation}
We compute the Pearson correlation between the transfer score and distance vector between each language pair.
The detailed results are presented figures \ref{fig:app_corr_zscls_m} and \ref{fig:app_corr_zscls_s} in appendix \ref{app:detail}.
While there are several distances that achieve a correlation score of over 0.7, effectively predicting the corresponding transfer performances, none of the linguistic relations are highly correlated with the transfer scores for both tasks.
In particular, syntax and inventory features have above-average correlation scores for SMiLER, whereas only phonology-based distances are effective for the event detection task.

\textbf{Combined Metric}
In order to achieve our objective of creating a universal metric that can be applied across different practical settings, we define a combined metric as a weighted average of all relevant linguistic features.
For each task, the optimal weights are the solution of a simple constrained correlation linear maximization (the weights are constrained to be non-negative and sum to 1).
Figure \ref{fig:LingFeat} compares the resulting weight importances between the two tasks across model scales.
Similar to the above assessment, there is a divergence between MED and SMilER on how the linguistic features are weighted in the optimal combined metric.

From these observations, we propose a joint combined metric that involves all three of the typological features as follows: $d_{comb} = 0.4 * d_\textit{ander-syntax} + 0.2 * d_\textit{inner-phonology} + 0.4 * d_\textit{ander-inventory}$.
To demonstrate the adaptability of the new distance, we provide the mean correlation scores (across all languages) of all computed linguistic distances in figure \ref{fig:CombCorr}.
Not only $d_{comb}$ achieves the highest correlation with transfer performances overall (above 0.6 for every setting), the combined metric also greatly lessens the score's variability amid tasks and scales of models.
This implies that $d_{comb}$ has the potential to be a general metric to approximate ZSCL performances prior to model training.
The following sections will use this combined distance for guiding the language selection and adversarial training in multi-transfer setting.






%

\begin{figure}[ht!]
\captionsetup{skip=2pt}
\addtolength{\belowcaptionskip}{-7mm}
\begin{center} 
\includegraphics[width=0.36\textwidth]{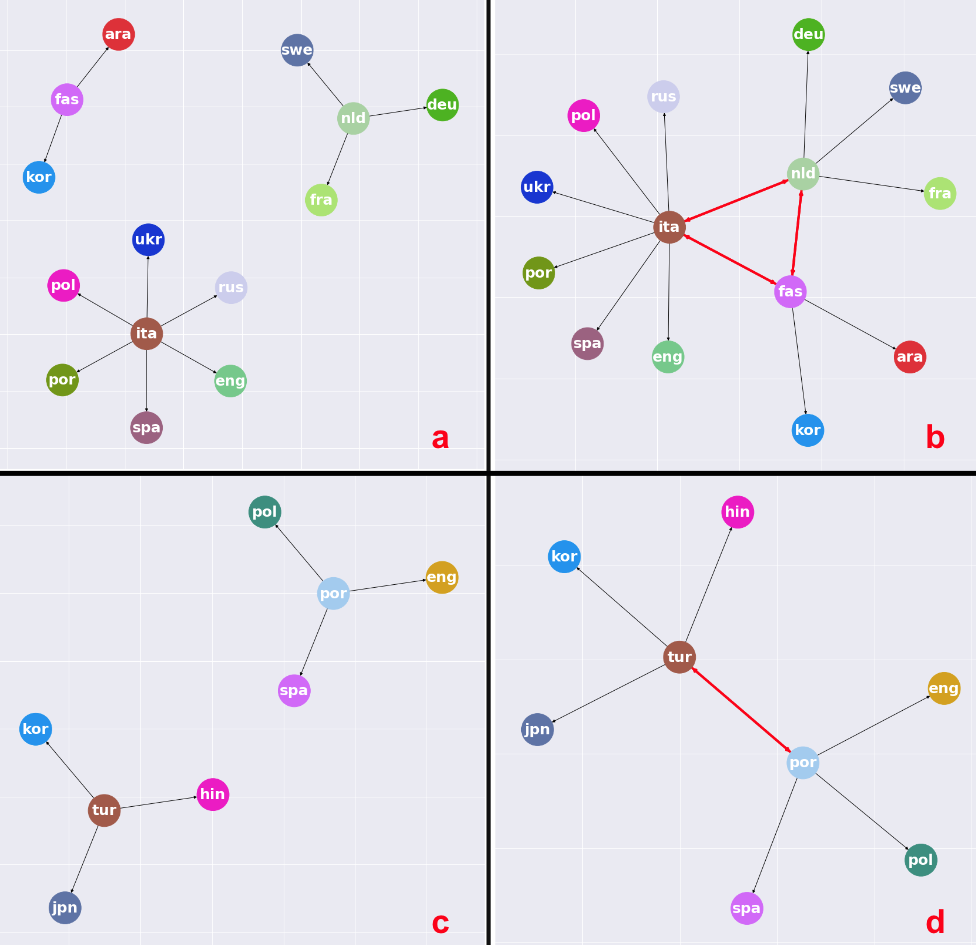}
\caption{\small Language clustering results for languages in SMiLER (a and b) and MINION (c and d). The graphs on the right (b and d) are the same as the ones on the left, but with connected medoids indicated by the new red edges.} \label{fig:LangGraph}
\end{center}
\end{figure}

\begin{table}[htbp]
  \centering
  \small
\addtolength{\belowcaptionskip}{-5mm}
\resizebox{0.75\columnwidth}{!}{
\begin{tabular}{cc|rrr|r}
\multicolumn{1}{l}{}                     & \multicolumn{1}{l|}{}             & \multicolumn{1}{l}{SMALL}   & \multicolumn{1}{l}{BASE}    & \multicolumn{1}{l|}{LARGE}  & \multicolumn{1}{l}{MODEL\_AVG} \\ \hline
\multicolumn{1}{c|}{}                    & \cellcolor[HTML]{FFFFFF}medoids*   & \cellcolor[HTML]{D9EAD3}1.8 & \cellcolor[HTML]{D9EAD3}1.7 & \cellcolor[HTML]{D9EAD3}0.7 & \cellcolor[HTML]{D9EAD3}1.4    \\
\multicolumn{1}{c|}{}                    & \cellcolor[HTML]{FFFFFF}tur*      & \cellcolor[HTML]{B6D7A8}2.7 & \cellcolor[HTML]{D9EAD3}1.0 & \cellcolor[HTML]{D9EAD3}1.0 & \cellcolor[HTML]{D9EAD3}1.5    \\
\multicolumn{1}{c|}{}                    & \cellcolor[HTML]{FFFFFF}por*      & \cellcolor[HTML]{93C47D}6.0 & \cellcolor[HTML]{93C47D}6.1 & \cellcolor[HTML]{93C47D}5.8 & \cellcolor[HTML]{93C47D}6.0    \\ \hhline{~|-----}
\multicolumn{1}{c|}{\multirow{-4}{*}{\small{M}}} & \cellcolor[HTML]{FFFFFF}task\_avg & \cellcolor[HTML]{B6D7A8}3.5 & \cellcolor[HTML]{B6D7A8}2.9 & \cellcolor[HTML]{B6D7A8}2.5 & \cellcolor[HTML]{B6D7A8}3.0    \\ \hline \hline
\multicolumn{1}{c|}{}                    & \cellcolor[HTML]{FFFFFF}medoids*   & \cellcolor[HTML]{B6D7A8}2.1 & \cellcolor[HTML]{D9EAD3}1.4 & \cellcolor[HTML]{D9EAD3}1.9 & \cellcolor[HTML]{D9EAD3}1.8    \\
\multicolumn{1}{c|}{}                    & \cellcolor[HTML]{FFFFFF}ita*      & \cellcolor[HTML]{B6D7A8}3.9 & \cellcolor[HTML]{B6D7A8}3.7 & \cellcolor[HTML]{B6D7A8}3.1 & \cellcolor[HTML]{B6D7A8}3.5    \\
\multicolumn{1}{c|}{}                    & \cellcolor[HTML]{FFFFFF}nld*      & \cellcolor[HTML]{93C47D}9.2 & \cellcolor[HTML]{93C47D}8.6 & \cellcolor[HTML]{93C47D}6.7 & \cellcolor[HTML]{93C47D}8.2    \\
\multicolumn{1}{c|}{}                    & \cellcolor[HTML]{FFFFFF}fas*      & \cellcolor[HTML]{D9EAD3}1.8 & \cellcolor[HTML]{D9EAD3}1.7 & \cellcolor[HTML]{D9EAD3}0.7 & \cellcolor[HTML]{D9EAD3}1.4    \\ \hhline{~|-----}
\multicolumn{1}{c|}{\multirow{-5}{*}{\small{S}}} & \cellcolor[HTML]{FFFFFF}task\_avg & \cellcolor[HTML]{93C47D}4.2 & \cellcolor[HTML]{B6D7A8}3.8 & \cellcolor[HTML]{B6D7A8}3.1 & \cellcolor[HTML]{B6D7A8}3.7   
\end{tabular}
}
\caption{\small{Differences in ZSCL-M scores (F1) of Inter-cluster (\textit{medoids*}) and Intra-cluster (\textit{medoid\_lang]*}) configurations over Random configuration, for tasks MED (M) and SMiLER (S).}}
\label{tab:zsclm}
\end{table}

\section{Zero-shot Cross-lingual Multi-transfer}
We address question Q2 by evaluating multi-transfer performances between two sets of languages.
In particular, we define a transfer configuration as an experimental setting that specifies languages inside the source and target sets, and a transfer run as an actual experimental evaluation of a transfer configuration.
As the number of configurations is exponential in terms of the number of languages, it is computationally impossible to evaluate every configuration, even more so for different tasks and model scales.
Therefore, we focus solely on the resource-constrained scenario, which is also equivalent to the setting with the minimum number of source languages.
Based on the result from the previous section, we propose to limit the configuration scope using the general combined distance metric as follows.

\subsection{Language Selection}
In the resource-constrained setting, our goal is to select the minimal set of source languages $D_s$ that can maximally transfer to a given target language set $D_t$.
We further restrict our attention to transfer configurations with $D_t$ as set of closely related languages in terms of cross-lingual transfer.
Assuming pair-wise transfer is highly correlated with multi-transferm, these configurations can be identified by clustering languages based on the combined linguistic distance $d_{comb}$.







\textbf{Language Clustering}
We use k-means \cite{kmeans} to partition the available languages into clusters based on $d_{comb}$.
In particular, k-medoids \footnote{\url{https://en.wikipedia.org/wiki/K-medoids}}, a variant of k-means is used for both clustering and finding the medoids (the actual data point/language that is the center of each cluster).
These medoid languages should be the optimal language that transfers best to every member of its cluster.
The resulting clusterings are shown in figure \ref{fig:LangGraph} for both MINION (\ref{fig:LangGraph}c) and SMiLER (\ref{fig:LangGraph}a).
The minimal number of source languages ($|D_s|$) is chosen to be equal to the minimal number of clusters such that each cluster has at least 3 members (so we can meaningfully evaluate multi-transfer setting).

\subsection{Experimental Setup}
In ZSCL-M, source training set $D_s$ consists of $N_s$ labeled datasets, and the goal is to transfer to target cluster $D_t$.
For MINION task, $N_s = 2$ and $D_t$ can be the turkish cluster $tur*$, or the portuguese cluster $por*$.
For SMiLER, $Ns = 3$ and $D_t$ can be the italian cluster $ita*$, the dutch cluster $nld*$, or the farsi cluster $fas*$
We are interested in verifying the following two hypotheses:

\textbf{(1) Inter-Cluster Transfer:} The best set of languages that transfer best across every cluster is the set of every medoid language. 

\textbf{(2) Intra-Cluster Transfer:} The best set of languages that transfer best for a given cluster is a subset of that cluster.

\subsection{Transfer Performance} \label{hypo}
Detailed results of multi-transfer performances are provided in figures \ref{fig:app_zsclm_m} and \ref{fig:app_zsclm_s} in appendix \ref{app:detail}, from which we can observe a clear improvement over single-transfer setting owning the additional training data.
Our main interest here is how effective $d_{comb}$ is in guiding the language selection for ZSCL-M.
Table. \ref{tab:zsclm} shows the differences in transfer scores of Inter-cluster ($medoids*$) and Intra-cluster ($[medoid\_lang]*$) configurations over Random configuration.
Specifically, $medoids*$ measures inter-cluster transfer capacity of the set $D_s$ consisting of every medoid from each cluster, to the target set $D_t$ of all considered languages.
On the other hand, $[medoid\_lang]*$ measures intra-cluster performance of a randomly sampled set $D_s$ of $N_s$ languages from the corresponding cluster, to the target set $D_t$ of every language in that cluster.
Finally, Random configurations are sampled from the set of configurations that are not part of the above two configurations.
Except for Inter-cluster configuration which only has one option, the results of Intra-cluster and Random configurations are the average of their sampled transfer runs.

The results from table \ref{tab:zsclm} show that language selection based on $d_{comb}$ provides a considerable boost in multi-transfer scores for every (task, configuration, model scale) setting.
This suggests that these medoid languages have the potential to achieve optimal cost-performance trade-offs in multi-transfer setting.
Notably between the two tasks, SMiLER has higher performance increases with only one additional source languages, despite having almost double the number of languages in target set.
This implies that as the number of languages grows considerably larger, there may exist a magnitude smaller set of optimal universal languages (medoids) that are able to transfer to every language extremely well.

 
\begin{table}[htbp]
  \centering
  \small
\addtolength{\belowcaptionskip}{-4mm}
\resizebox{\columnwidth}{!}{
\begin{tabular}{cc|cc|cc|cc|cc}
\multicolumn{1}{l}{} & \multicolumn{1}{l|}{} & \multicolumn{2}{c|}{SMALL} & \multicolumn{2}{c|}{BASE} & \multicolumn{2}{c|}{LARGE} & \multicolumn{2}{c}{MODEL\_AVG} \\ \cline{3-10} 
\multicolumn{1}{l}{} & \multicolumn{1}{l|}{} & ZSCL-R & DANN & ZSCL-R & DANN & ZSCL-R & DANN & ZSCL-R & DANN \\ \hline
\multicolumn{1}{c|}{} & \cellcolor[HTML]{FFFFFF}medoid* & \cellcolor[HTML]{D9EAD3}0.7 & \cellcolor[HTML]{EA9999}-3.7 & \cellcolor[HTML]{D9EAD3}1.4 & \cellcolor[HTML]{F4CCCC}-1.9 & \cellcolor[HTML]{D9EAD3}1.3 & \cellcolor[HTML]{F4CCCC}-2.8 & \cellcolor[HTML]{D9EAD3}1.1 & \cellcolor[HTML]{F4CCCC}-2.8 \\
\multicolumn{1}{c|}{} & \cellcolor[HTML]{FFFFFF}tur* & \cellcolor[HTML]{93C47D}4.1 & \cellcolor[HTML]{FCE5CD}-0.5 & \cellcolor[HTML]{B6D7A8}2.7 & \cellcolor[HTML]{F4CCCC}-2.7 & \cellcolor[HTML]{93C47D}3.6 & \cellcolor[HTML]{FCE5CD}-0.1 & \cellcolor[HTML]{93C47D}3.5 & \cellcolor[HTML]{F4CCCC}-1.1 \\
\multicolumn{1}{c|}{} & \cellcolor[HTML]{FFFFFF}por* & \cellcolor[HTML]{D9EAD3}0.6 & \cellcolor[HTML]{EA9999}-5.5 & \cellcolor[HTML]{FCE5CD}-0.3 & \cellcolor[HTML]{EA9999}-4.2 & \cellcolor[HTML]{FCE5CD}-0.3 & \cellcolor[HTML]{EA9999}-5.8 & \cellcolor[HTML]{FFFFFF}0.0 & \cellcolor[HTML]{EA9999}-5.2 \\  \hhline{~|---------}
\multicolumn{1}{c|}{\multirow{-4}{*}{M}} & \cellcolor[HTML]{FFFFFF}task\_avg & \cellcolor[HTML]{B6D7A8}1.8 & \cellcolor[HTML]{EA9999}-3.2 & \cellcolor[HTML]{D9EAD3}1.3 & \cellcolor[HTML]{F4CCCC}-2.9 & \cellcolor[HTML]{B6D7A8}1.5 & \cellcolor[HTML]{F4CCCC}-2.9 & \cellcolor[HTML]{D9EAD3}1.5 & \cellcolor[HTML]{EA9999}-3.0 \\ \hline \hline
\multicolumn{1}{c|}{} & \cellcolor[HTML]{FFFFFF}medoid* & \cellcolor[HTML]{B6D7A8}1.8 & \cellcolor[HTML]{E06666}-11.8 & \cellcolor[HTML]{93C47D}3.2 & \cellcolor[HTML]{EA9999}-4.2 & \cellcolor[HTML]{D9EAD3}0.6 & \cellcolor[HTML]{F4CCCC}-1.7 & \cellcolor[HTML]{D9EAD3}1.9 & \cellcolor[HTML]{EA9999}-5.9 \\
\multicolumn{1}{c|}{} & \cellcolor[HTML]{FFFFFF}ita* & \cellcolor[HTML]{93C47D}3.0 & \cellcolor[HTML]{E06666}-15.1 & \cellcolor[HTML]{93C47D}3.9 & \cellcolor[HTML]{EA9999}-5.5 & \cellcolor[HTML]{93C47D}2.3 & \cellcolor[HTML]{EA9999}-5.8 & \cellcolor[HTML]{93C47D}3.1 & \cellcolor[HTML]{EA9999}-8.8 \\
\multicolumn{1}{c|}{} & \cellcolor[HTML]{FFFFFF}nld* & \cellcolor[HTML]{93C47D}2.4 & \cellcolor[HTML]{E06666}-10.9 & \cellcolor[HTML]{93C47D}2.8 & \cellcolor[HTML]{EA9999}-4.9 & \cellcolor[HTML]{FFFFFF}0.0 & \cellcolor[HTML]{EA9999}-5.1 & \cellcolor[HTML]{B6D7A8}1.7 & \cellcolor[HTML]{EA9999}-7.0 \\
\multicolumn{1}{c|}{} & \cellcolor[HTML]{FFFFFF}fas* & \cellcolor[HTML]{FFF2CC}-0.2 & \cellcolor[HTML]{E06666}-10.6 & \cellcolor[HTML]{93C47D}2.1 & \cellcolor[HTML]{EA9999}-4.2 & \cellcolor[HTML]{93C47D}2.4 & \cellcolor[HTML]{F4CCCC}-2.7 & \cellcolor[HTML]{B6D7A8}1.4 & \cellcolor[HTML]{EA9999}-5.8 \\  \hhline{~|---------}
\multicolumn{1}{c|}{\multirow{-5}{*}{S}} & \cellcolor[HTML]{FFFFFF}task\_avg & \cellcolor[HTML]{B6D7A8}1.8 & \cellcolor[HTML]{E06666}-12.1 & \cellcolor[HTML]{93C47D}3.0 & \cellcolor[HTML]{EA9999}-4.7 & \cellcolor[HTML]{B6D7A8}1.3 & \cellcolor[HTML]{F4CCCC}-3.8 & \cellcolor[HTML]{93C47D}2.0 & \cellcolor[HTML]{EA9999}-6.9
\end{tabular}
}
\caption{Difference between transfer performances of adversarial training methods and ZSCL-M runs in inter-cluster setting.}
\label{tab:zsclr}
\end{table}

\section{Zero-shot Cross-lingual Relational-transfer}
Due to limited access to multi-lingual annotators, gathering labeled data across languages is difficult.
Previous section address this by careful language/data selection to optimize cost-effect.
In contrast, unlabeled data is easy to collect, but leveraging it correctly for ZSCL is non-trivial.
This section investigates the effectiveness of adversarial training approach for the more general ZSCL-M setting, and the possibility of further improving multi-lingual transfer through explicitly integrating our transfer-correlated linguistic relations.


\subsection{Experimental Setup}
We follow the same setup as in ZSCL-M, but each language is accompanied by an unlabeled dataset which can be used for training.
The model use labeled data from the source cluster to learn the task, whereas unlabeled data from another cluster is used to help transfer source performance to that target cluster.
The goal is to bridge the performance between 2 different language clusters with the aid of given unlabeled text.

\textbf{Adversarial Language Adaptation} A typical method use for ZSCL-S is adversarial language adaptation (ALA) \cite{sentiment, nguyen21b} which employs a language discriminator that takes an encoded representation from mLMs as its input and predicts its origin (language).
By pushing the encoder to both minimize the downstream loss and maximally misdirect the language predictor (adversarial training), the resulting representation can be indiscriminate with respect to the shift between the languages while also discriminative for the main learning task.
Apply ALA for ZSCL-M setting is equivalent to applying DANN for a single joint source domain to a single joint target domain (the union of every languages in $D_s$ and $D_t$, respectively).
 
\textbf{Zero-shot Cross-lingual Relational-transfer} We extend ALA to the case of multiple source and target languages through Graph-relational domain adaptation (GrDA) \cite{grda}, a generalization of DANN to multi-domains adaptation setting by introducing a domain graph that captures heterogeneous relations among domains.
GrDA relaxes the strict uniform alignment of DANN to allow flexible and effective adaptation between distant domains.
We use the language clustering graphs on the right of figure \ref{fig:LangGraph} as domain graphs for GrDA.
Noted that additional direct connections between medoid languages are introduced (red edges) to facilitate inter-cluster transfer.
We refer to this adversarial learning process for ZSCL that directly embeds the linguistic relations into the representations as Zero-shot Cross-lingual Relational-transfer (ZSCL-R).

\subsection{Transfer Performance}
Performances of the baseline DANN and the proposed adversarial training method ZSCL-R are provided in table. \ref{tab:zsclr}, which follows the same format as table. \ref{tab:zsclm}.
However, instead of comparing against Random configurations, they are compared directly with results of ZSCL-M runs of the corresponding configurations, but only in terms of inter-cluster transfer.
The negative results of DANN confirm that strictly aligning language representations uniformly is not effective in  ZSCL-M setting.
As the model scale gets smaller, model's representation becomes less expressive, whereas the language-invariant feature of all languages is harder to capture as the number of languages grows. Thus, the adverse effect gets significantly worse on small models for SMiLER task (-12 points on average).
In contrast, ZSCL-R provides consistent improvements over ZSCL-M for most configurations.
Due to the flexibility of GrDA alignment, ZSCL-R performs even better as the number of languages increases, effectively leveraging the additional unlabeled data to help improve inter-cluster transfer ability of models.




\section{Conclusion}
We explored the general cross-lingual transfer learning setting where multiple source and target languages are involved.
Our experiments on two practical information extraction tasks across different model scales and languages reveal new general insights on cross-lingual transfer learning:
(1) There is a correlation between linguistic distances and single-transfer performances; however, simplistic measures based on syntax features are only sufficient for syntactic-based tasks.
We develop a combined distance based on various metrics and linguistic features that achieves a high correlation with cross-lingual transfer score robustly across all settings.
(2) The proposed combined metric provides useful directions for language clustering and selection to achieve optimal cost-performance trade-off in multi-transfer to a specific group of languages.
(3) Finally, linguistic relations can be leveraged with unlabeled data for adversarial training to help generalize to a new group of languages with minimum additional annotation cost.
Our findings collectively suggest multi-transfer as a new baseline for cross-lingual learning, and provide a baseline for efficient and effective multi-transfer together with promising directions that future work can further improve upon.





\section*{Limitations}
Compared to prior cross-lingual transfer papers \cite{litmus}, our work aims to demonstrate generalization across various hyperparameters and design choices that affect the results of previous investigations on the topic.
Consequently, this has led to significant computational demands, forcing us to limit and simplify some aspects of our experiments to ensure manageability.
This section outlines what we have and has not been able to addressed, and suggests promising future directions that can be followed from our findings.

First, our combined metric is heuristically defined based on the transfer-distance correlation scores.
Although the optimal metric for all situations is impossible to find and likely non-existent, we anticipate that the ideal metric won't differ substantially across settings of tasks and model architectures, and may only vary slightly from ours. Nevertheless, an in-depth analysis is needed on how a significant change in the distance metric can impact ZSML-M and particularly ZSML-R, which is explicitly guided by the metric.

Second, We only experiment with a minimum number of language clusters needed for an effective evaluation of multi-transfer.
There is no guarantee that this minimal cost strategy is also the one with the best cost-performance trade-off, which can depend on the number of languages and task availability.
Future work may investigate this trade-off as the problem scales to hundreds of languages, in particular validating the hypothesis stated in section \ref{hypo}: the best trade-off point (the number of clusters) is an order of magnitude smaller than the number of available languages.

Third, the decision to connect the medoids in the language cluster graph for ZSCL-R is the simplest solution to create a connected language-relation graph. This, however, also implies the path between any two languages has a maximum length of 3, which does not reflect the actual relation distance among languages.
Further research is needed to ascertain the optimal language relational-graph, especially in intricate scenarios that involve many languages and tasks.

Finally, our model scale only stops at hundreds of millions of parameters, which are no longer considered large scale by today's standard.
Further experimentation can test if our results hold for the current billion scale models.
A more interesting direction would be to investigate cross-lingual multi-transfer performance of parameter-efficient tuning \cite{chen-etal-2022-revisiting} and instruction tuning \cite{instruct}, which have become the standard approaches for fine-tuning these massive scale models.



\bibliography{anthology,custom}

\begin{thebibliography}{38}
\expandafter\ifx\csname natexlab\endcsname\relax\def\natexlab#1{#1}\fi

\bibitem[{Ammar et~al.(2016)Ammar, Mulcaire, Ballesteros, Dyer, and
  Smith}]{ammar16}
Waleed Ammar, George Mulcaire, Miguel Ballesteros, Chris Dyer, and Noah~A.
  Smith. 2016.
\newblock \href {https://doi.org/10.1162/tacl_a_00109} {Many languages, one
  parser}.
\newblock \emph{Transactions of the Association for Computational Linguistics},
  4:431--444.

\bibitem[{Chen et~al.(2022)Chen, Liu, Meng, and
  Liang}]{chen-etal-2022-revisiting}
Guanzheng Chen, Fangyu Liu, Zaiqiao Meng, and Shangsong Liang. 2022.
\newblock \href {https://aclanthology.org/2022.emnlp-main.168} {Revisiting
  parameter-efficient tuning: Are we really there yet?}
\newblock In \emph{Proceedings of the 2022 Conference on Empirical Methods in
  Natural Language Processing}, pages 2612--2626, Abu Dhabi, United Arab
  Emirates. Association for Computational Linguistics.

\bibitem[{Chen et~al.(2018)Chen, Sun, Athiwaratkun, Cardie, and
  Weinberger}]{sentiment}
Xilun Chen, Yu~Sun, Ben Athiwaratkun, Claire Cardie, and Kilian Weinberger.
  2018.
\newblock \href {https://doi.org/10.1162/tacl_a_00039} {Adversarial deep
  averaging networks for cross-lingual sentiment classification}.
\newblock \emph{Transactions of the Association for Computational Linguistics},
  6:557--570.

\bibitem[{Chi et~al.(2021)Chi, Dong, Wei, Yang, Singhal, Wang, Song, Mao,
  Huang, and Zhou}]{infoxlm}
Zewen Chi, Li~Dong, Furu Wei, Nan Yang, Saksham Singhal, Wenhui Wang, Xia Song,
  Xian-Ling Mao, Heyan Huang, and Ming Zhou. 2021.
\newblock \href {https://doi.org/10.18653/v1/2021.naacl-main.280} {{InfoXLM}:
  An information-theoretic framework for cross-lingual language model
  pre-training}.
\newblock In \emph{Proceedings of the 2021 Conference of the North American
  Chapter of the Association for Computational Linguistics: Human Language
  Technologies}. Association for Computational Linguistics.

\bibitem[{Choi et~al.(2009)Choi, Cha, and Tappert}]{binary}
SHC Choi, Sung-Hyuk Cha, and Charles Tappert. 2009.
\newblock A survey of binary similarity and distance measures.
\newblock \emph{J. Syst. Cybern. Inf.}, 8.

\bibitem[{Conneau et~al.(2020)Conneau, Khandelwal, Goyal, Chaudhary, Wenzek,
  Guzm{\'a}n, Grave, Ott, Zettlemoyer, and Stoyanov}]{xlmr}
Alexis Conneau, Kartikay Khandelwal, Naman Goyal, Vishrav Chaudhary, Guillaume
  Wenzek, Francisco Guzm{\'a}n, Edouard Grave, Myle Ott, Luke Zettlemoyer, and
  Veselin Stoyanov. 2020.
\newblock \href {https://doi.org/10.18653/v1/2020.acl-main.747} {Unsupervised
  cross-lingual representation learning at scale}.
\newblock In \emph{Proceedings of the 58th Annual Meeting of the Association
  for Computational Linguistics}, pages 8440--8451, Online. Association for
  Computational Linguistics.

\bibitem[{Dolicki and Spanakis(2021)}]{dolicki:21}
Błażej Dolicki and Gerasimos Spanakis. 2021.
\newblock \href {http://arxiv.org/abs/2105.05975} {Analysing the impact of
  linguistic features on cross-lingual transfer}.

\bibitem[{Dryer and Haspelmath(2013)}]{wals}
Matthew~S. Dryer and Martin Haspelmath, editors. 2013.
\newblock \href {https://doi.org/10.5281/zenodo.7385533} {\emph{WALS Online
  (v2020.3)}}.
\newblock Zenodo.

\bibitem[{Fang et~al.(2021)Fang, Wang, Gan, Sun, and Liu}]{filter}
Yuwei Fang, Shuohang Wang, Zhe Gan, Siqi Sun, and Jingjing Liu. 2021.
\newblock \href {https://doi.org/10.1609/aaai.v35i14.17512} {{FILTER}: An
  enhanced fusion method for cross-lingual language understanding}.
\newblock \emph{Proceedings of the {AAAI} Conference on Artificial
  Intelligence}, 35(14):12776--12784.

\bibitem[{Ganin et~al.(2016)Ganin, Ustinova, Ajakan, Germain, Larochelle,
  Laviolette, March, and Lempitsky}]{dann}
Yaroslav Ganin, Evgeniya Ustinova, Hana Ajakan, Pascal Germain, Hugo
  Larochelle, Fran{\c{c}}ois Laviolette, Mario March, and Victor Lempitsky.
  2016.
\newblock \href {http://jmlr.org/papers/v17/15-239.html} {Domain-adversarial
  training of neural networks}.
\newblock \emph{Journal of Machine Learning Research}, 17(59):1--35.

\bibitem[{Goyal et~al.(2021)Goyal, Du, Ott, Anantharaman, and
  Conneau}]{xlmr-xl}
Naman Goyal, Jingfei Du, Myle Ott, Giri Anantharaman, and Alexis Conneau. 2021.
\newblock \href {https://doi.org/10.18653/v1/2021.repl4nlp-1.4} {Larger-scale
  transformers for multilingual masked language modeling}.
\newblock In \emph{Proceedings of the 6th Workshop on Representation Learning
  for {NLP} ({RepL}4NLP-2021)}. Association for Computational Linguistics.

\bibitem[{Hammarström et~al.(2022)Hammarström, Forkel, Haspelmath, and
  Bank}]{glottolog}
Harald Hammarström, Robert Forkel, Martin Haspelmath, and Sebastian Bank.
  2022.
\newblock \href {https://doi.org/10.5281/zenodo.7398962} {glottolog/glottolog:
  Glottolog database 4.7}.

\bibitem[{Huang et~al.(2019)Huang, Ji, and May}]{nametag}
Lifu Huang, Heng Ji, and Jonathan May. 2019.
\newblock \href {https://doi.org/10.18653/v1/N19-1383} {Cross-lingual
  multi-level adversarial transfer to enhance low-resource name tagging}.
\newblock In \emph{Proceedings of the 2019 Conference of the North {A}merican
  Chapter of the Association for Computational Linguistics: Human Language
  Technologies, Volume 1 (Long and Short Papers)}, pages 3823--3833,
  Minneapolis, Minnesota. Association for Computational Linguistics.

\bibitem[{Keung et~al.(2020)Keung, Lu, Salazar, and Bhardwaj}]{keung20}
Phillip Keung, Yichao Lu, Julian Salazar, and Vikas Bhardwaj. 2020.
\newblock \href {https://doi.org/10.18653/v1/2020.emnlp-main.40} {Don{'}t use
  {E}nglish dev: On the zero-shot cross-lingual evaluation of contextual
  embeddings}.
\newblock In \emph{Proceedings of the 2020 Conference on Empirical Methods in
  Natural Language Processing (EMNLP)}, pages 549--554, Online. Association for
  Computational Linguistics.

\bibitem[{Lauscher et~al.(2020)Lauscher, Ravishankar, Vuli{\'c}, and
  Glava{\v{s}}}]{hero}
Anne Lauscher, Vinit Ravishankar, Ivan Vuli{\'c}, and Goran Glava{\v{s}}. 2020.
\newblock \href {https://doi.org/10.18653/v1/2020.emnlp-main.363} {From zero to
  hero: {O}n the limitations of zero-shot language transfer with multilingual
  {T}ransformers}.
\newblock In \emph{Proceedings of the 2020 Conference on Empirical Methods in
  Natural Language Processing (EMNLP)}, pages 4483--4499, Online. Association
  for Computational Linguistics.

\bibitem[{Lewis(2009)}]{ethnologue}
M.~Paul Lewis, editor. 2009.
\newblock \emph{Ethnologue: Languages of the World}, sixteenth edition.
\newblock SIL International, Dallas, TX, USA.

\bibitem[{Liang et~al.(2020)Liang, Duan, Gong, Wu, Guo, Qi, Gong, Shou, Jiang,
  Cao, Fan, Zhang, Agrawal, Cui, Wei, Bharti, Qiao, Chen, Wu, Liu, Yang,
  Campos, Majumder, and Zhou}]{xglue}
Yaobo Liang, Nan Duan, Yeyun Gong, Ning Wu, Fenfei Guo, Weizhen Qi, Ming Gong,
  Linjun Shou, Daxin Jiang, Guihong Cao, Xiaodong Fan, Ruofei Zhang, Rahul
  Agrawal, Edward Cui, Sining Wei, Taroon Bharti, Ying Qiao, Jiun-Hung Chen,
  Winnie Wu, Shuguang Liu, Fan Yang, Daniel Campos, Rangan Majumder, and Ming
  Zhou. 2020.
\newblock \href {https://doi.org/10.18653/v1/2020.emnlp-main.484} {{XGLUE}: A
  new benchmark datasetfor cross-lingual pre-training, understanding and
  generation}.
\newblock In \emph{Proceedings of the 2020 Conference on Empirical Methods in
  Natural Language Processing ({EMNLP})}. Association for Computational
  Linguistics.

\bibitem[{Limisiewicz et~al.(2020)Limisiewicz, Mare{\v{c}}ek, and
  Rosa}]{limisiewicz20}
Tomasz Limisiewicz, David Mare{\v{c}}ek, and Rudolf Rosa. 2020.
\newblock \href {https://doi.org/10.18653/v1/2020.findings-emnlp.245}
  {{U}niversal {D}ependencies {A}ccording to {BERT}: {B}oth {M}ore {S}pecific
  and {M}ore {G}eneral}.
\newblock In \emph{Findings of the Association for Computational Linguistics:
  EMNLP 2020}, pages 2710--2722, Online. Association for Computational
  Linguistics.

\bibitem[{Lin et~al.(2019)Lin, Chen, Lee, Li, Zhang, Xia, Rijhwani, He, Zhang,
  Ma, Anastasopoulos, Littell, and Neubig}]{lin:19}
Yu-Hsiang Lin, Chian-Yu Chen, Jean Lee, Zirui Li, Yuyan Zhang, Mengzhou Xia,
  Shruti Rijhwani, Junxian He, Zhisong Zhang, Xuezhe Ma, Antonios
  Anastasopoulos, Patrick Littell, and Graham Neubig. 2019.
\newblock \href {https://doi.org/10.18653/v1/P19-1301} {Choosing transfer
  languages for cross-lingual learning}.
\newblock In \emph{Proceedings of the 57th Annual Meeting of the Association
  for Computational Linguistics}, pages 3125--3135, Florence, Italy.
  Association for Computational Linguistics.

\bibitem[{Littell et~al.(2017)Littell, Mortensen, Lin, Kairis, Turner, and
  Levin}]{uriel}
Patrick Littell, David~R. Mortensen, Ke~Lin, Katherine Kairis, Carlisle Turner,
  and Lori Levin. 2017.
\newblock \href {https://aclanthology.org/E17-2002} {{URIEL} and lang2vec:
  Representing languages as typological, geographical, and phylogenetic
  vectors}.
\newblock In \emph{Proceedings of the 15th Conference of the {E}uropean Chapter
  of the Association for Computational Linguistics: Volume 2, Short Papers},
  pages 8--14, Valencia, Spain. Association for Computational Linguistics.

\bibitem[{Lloyd(1982)}]{kmeans}
S.~Lloyd. 1982.
\newblock \href {https://doi.org/10.1109/TIT.1982.1056489} {Least squares
  quantization in pcm}.
\newblock \emph{IEEE Transactions on Information Theory}, 28(2):129--137.

\bibitem[{Malkin et~al.(2022)Malkin, Limisiewicz, and Stanovsky}]{malkin:22}
Dan Malkin, Tomasz Limisiewicz, and Gabriel Stanovsky. 2022.
\newblock \href {https://doi.org/10.18653/v1/2022.naacl-main.361} {A balanced
  data approach for evaluating cross-lingual transfer: Mapping the linguistic
  blood bank}.
\newblock In \emph{Proceedings of the 2022 Conference of the North American
  Chapter of the Association for Computational Linguistics: Human Language
  Technologies}. Association for Computational Linguistics.

\bibitem[{Moran et~al.(2014)Moran, McCloy, and Wright}]{phoible}
Steven Moran, Daniel McCloy, and Richard Wright, editors. 2014.
\newblock \href {http://phoible.org/} {\emph{PHOIBLE Online}}.
\newblock Max Planck Institute for Evolutionary Anthropology, Leipzig.

\bibitem[{Ngo~Trung et~al.(2021)Ngo~Trung, Phung, and Nguyen}]{ngo21}
Nghia Ngo~Trung, Duy Phung, and Thien~Huu Nguyen. 2021.
\newblock \href {https://doi.org/10.18653/v1/2021.findings-acl.351}
  {Unsupervised domain adaptation for event detection using domain-specific
  adapters}.
\newblock In \emph{Findings of the Association for Computational Linguistics:
  ACL-IJCNLP 2021}, pages 4015--4025, Online. Association for Computational
  Linguistics.

\bibitem[{Nguyen et~al.(2021)Nguyen, Nguyen, Min, and Nguyen}]{nguyen21b}
Minh~Van Nguyen, Tuan~Ngo Nguyen, Bonan Min, and Thien~Huu Nguyen. 2021.
\newblock \href {https://doi.org/10.18653/v1/2021.emnlp-main.440} {Crosslingual
  transfer learning for relation and event extraction via word category and
  class alignments}.
\newblock In \emph{Proceedings of the 2021 Conference on Empirical Methods in
  Natural Language Processing}, pages 5414--5426, Online and Punta Cana,
  Dominican Republic. Association for Computational Linguistics.

\bibitem[{Phang et~al.(2020)Phang, Calixto, Htut, Pruksachatkun, Liu, Vania,
  Kann, and Bowman}]{phang20}
Jason Phang, Iacer Calixto, Phu~Mon Htut, Yada Pruksachatkun, Haokun Liu, Clara
  Vania, Katharina Kann, and Samuel~R. Bowman. 2020.
\newblock \href {https://aclanthology.org/2020.aacl-main.56} {{E}nglish
  intermediate-task training improves zero-shot cross-lingual transfer too}.
\newblock In \emph{Proceedings of the 1st Conference of the Asia-Pacific
  Chapter of the Association for Computational Linguistics and the 10th
  International Joint Conference on Natural Language Processing}, pages
  557--575, Suzhou, China. Association for Computational Linguistics.

\bibitem[{Pires et~al.(2019)Pires, Schlinger, and Garrette}]{pires19}
Telmo Pires, Eva Schlinger, and Dan Garrette. 2019.
\newblock \href {https://doi.org/10.18653/v1/P19-1493} {How multilingual is
  multilingual {BERT}?}
\newblock In \emph{Proceedings of the 57th Annual Meeting of the Association
  for Computational Linguistics}, pages 4996--5001, Florence, Italy.
  Association for Computational Linguistics.

\bibitem[{Ponti et~al.(2018)Ponti, Reichart, Korhonen, and Vuli{\'c}}]{ponti18}
Edoardo~Maria Ponti, Roi Reichart, Anna Korhonen, and Ivan Vuli{\'c}. 2018.
\newblock \href {https://doi.org/10.18653/v1/P18-1142} {Isomorphic transfer of
  syntactic structures in cross-lingual {NLP}}.
\newblock In \emph{Proceedings of the 56th Annual Meeting of the Association
  for Computational Linguistics (Volume 1: Long Papers)}, pages 1531--1542,
  Melbourne, Australia. Association for Computational Linguistics.

\bibitem[{Pouran Ben~Veyseh et~al.(2022)Pouran Ben~Veyseh, Nguyen, Dernoncourt,
  and Nguyen}]{minion:22}
Amir Pouran Ben~Veyseh, Minh~Van Nguyen, Franck Dernoncourt, and Thien Nguyen.
  2022.
\newblock \href {https://doi.org/10.18653/v1/2022.naacl-main.166} {{MINION}: a
  large-scale and diverse dataset for multilingual event detection}.
\newblock In \emph{Proceedings of the 2022 Conference of the North American
  Chapter of the Association for Computational Linguistics: Human Language
  Technologies}, pages 2286--2299, Seattle, United States. Association for
  Computational Linguistics.

\bibitem[{Ruder et~al.(2021)Ruder, Constant, Botha, Siddhant, Firat, Fu, Liu,
  Hu, Garrette, Neubig, and Johnson}]{xtr-21}
Sebastian Ruder, Noah Constant, Jan Botha, Aditya Siddhant, Orhan Firat, Jinlan
  Fu, Pengfei Liu, Junjie Hu, Dan Garrette, Graham Neubig, and Melvin Johnson.
  2021.
\newblock \href {https://doi.org/10.18653/v1/2021.emnlp-main.802} {Xtreme-r:
  Towards more challenging and nuanced multilingual evaluation}.
\newblock \emph{Proceedings of the 2021 Conference on Empirical Methods in
  Natural Language Processing}.

\bibitem[{Seganti et~al.(2021)Seganti, Firl{\k{a}}g, Skowronska, Sat{\l}awa,
  and Andruszkiewicz}]{smiler:21}
Alessandro Seganti, Klaudia Firl{\k{a}}g, Helena Skowronska, Micha{\l}
  Sat{\l}awa, and Piotr Andruszkiewicz. 2021.
\newblock \href {https://doi.org/10.18653/v1/2021.eacl-main.166} {Multilingual
  entity and relation extraction dataset and model}.
\newblock In \emph{Proceedings of the 16th Conference of the European Chapter
  of the Association for Computational Linguistics: Main Volume}, pages
  1946--1955, Online. Association for Computational Linguistics.

\bibitem[{Srinivasan et~al.(2022)Srinivasan, Kholkar, Kejriwal, Ganu, Dandapat,
  Sitaram, Santhanam, Aditya, Bali, and Choudhury}]{litmus}
Anirudh Srinivasan, Gauri Kholkar, Rahul Kejriwal, Tanuja Ganu, Sandipan
  Dandapat, Sunayana Sitaram, Balakrishnan Santhanam, Somak Aditya, Kalika
  Bali, and Monojit Choudhury. 2022.
\newblock \href
  {https://www.microsoft.com/en-us/research/publication/litmus-predictor-an-ai-assistant-for-building-reliable-high-performing-and-fair-multilingual-nlp-systems/}
  {Litmus predictor: An ai assistant for building reliable, high-performing and
  fair multilingual nlp systems}.
\newblock In \emph{Thirty-sixth AAAI Conference on Artificial Intelligence}.
  AAAI.
\newblock System Demonstration.

\bibitem[{Srinivasan et~al.(2021)Srinivasan, Sitaram, Ganu, Dandapat, Bali, and
  Choudhury}]{srinivasan:21}
Anirudh Srinivasan, Sunayana Sitaram, Tanuja Ganu, Sandipan Dandapat, Kalika
  Bali, and Monojit Choudhury. 2021.
\newblock Predicting the performance of multilingual nlp models.
\newblock \emph{ArXiv}, abs/2110.08875.

\bibitem[{Turc et~al.(2021)Turc, Lee, Eisenstein, Chang, and
  Toutanova}]{primacy}
Iulia Turc, Kenton Lee, Jacob Eisenstein, Ming-Wei Chang, and Kristina
  Toutanova. 2021.
\newblock \href {http://arxiv.org/abs/2106.16171} {Revisiting the primacy of
  english in zero-shot cross-lingual transfer}.

\bibitem[{Wang et~al.(2020)Wang, Wei, Dong, Bao, Yang, and Zhou}]{minilm}
Wenhui Wang, Furu Wei, Li~Dong, Hangbo Bao, Nan Yang, and Ming Zhou. 2020.
\newblock Minilm: Deep self-attention distillation for task-agnostic
  compression of pre-trained transformers.
\newblock In \emph{Proceedings of the 34th International Conference on Neural
  Information Processing Systems}, NIPS'20, Red Hook, NY, USA. Curran
  Associates Inc.

\bibitem[{Wei et~al.(2022)Wei, Bosma, Zhao, Guu, Yu, Lester, Du, Dai, and
  Le}]{instruct}
Jason Wei, Maarten Bosma, Vincent Zhao, Kelvin Guu, Adams~Wei Yu, Brian Lester,
  Nan Du, Andrew~M. Dai, and Quoc~V Le. 2022.
\newblock \href {https://openreview.net/forum?id=gEZrGCozdqR} {Finetuned
  language models are zero-shot learners}.
\newblock In \emph{International Conference on Learning Representations}.

\bibitem[{Xu et~al.(2022{\natexlab{a}})Xu, Gui, Ma, Zhang, Ye, Zhang, and
  Huang}]{ningyu22}
Ningyu Xu, Tao Gui, Ruotian Ma, Qi~Zhang, Jingting Ye, Menghan Zhang, and
  Xuanjing Huang. 2022{\natexlab{a}}.
\newblock \href {https://aclanthology.org/2022.emnlp-main.552}
  {Cross-linguistic syntactic difference in multilingual {BERT}: How good is it
  and how does it affect transfer?}
\newblock In \emph{Proceedings of the 2022 Conference on Empirical Methods in
  Natural Language Processing}, pages 8073--8092, Abu Dhabi, United Arab
  Emirates. Association for Computational Linguistics.

\bibitem[{Xu et~al.(2022{\natexlab{b}})Xu, He, Lee, Wang, and Wang}]{grda}
Zihao Xu, Hao He, Guang-He Lee, Bernie Wang, and Hao Wang. 2022{\natexlab{b}}.
\newblock \href {https://openreview.net/forum?id=kcwyXtt7yDJ} {Graph-relational
  domain adaptation}.
\newblock In \emph{International Conference on Learning Representations}.

\end{thebibliography}
\bibliographystyle{acl_natbib}

\clearpage

\appendix

\section{Detailed Experimental Results} \label{app:detail}
\begin{sidewaysfigure*}
  \centering
  \includegraphics[width=0.95\textheight]{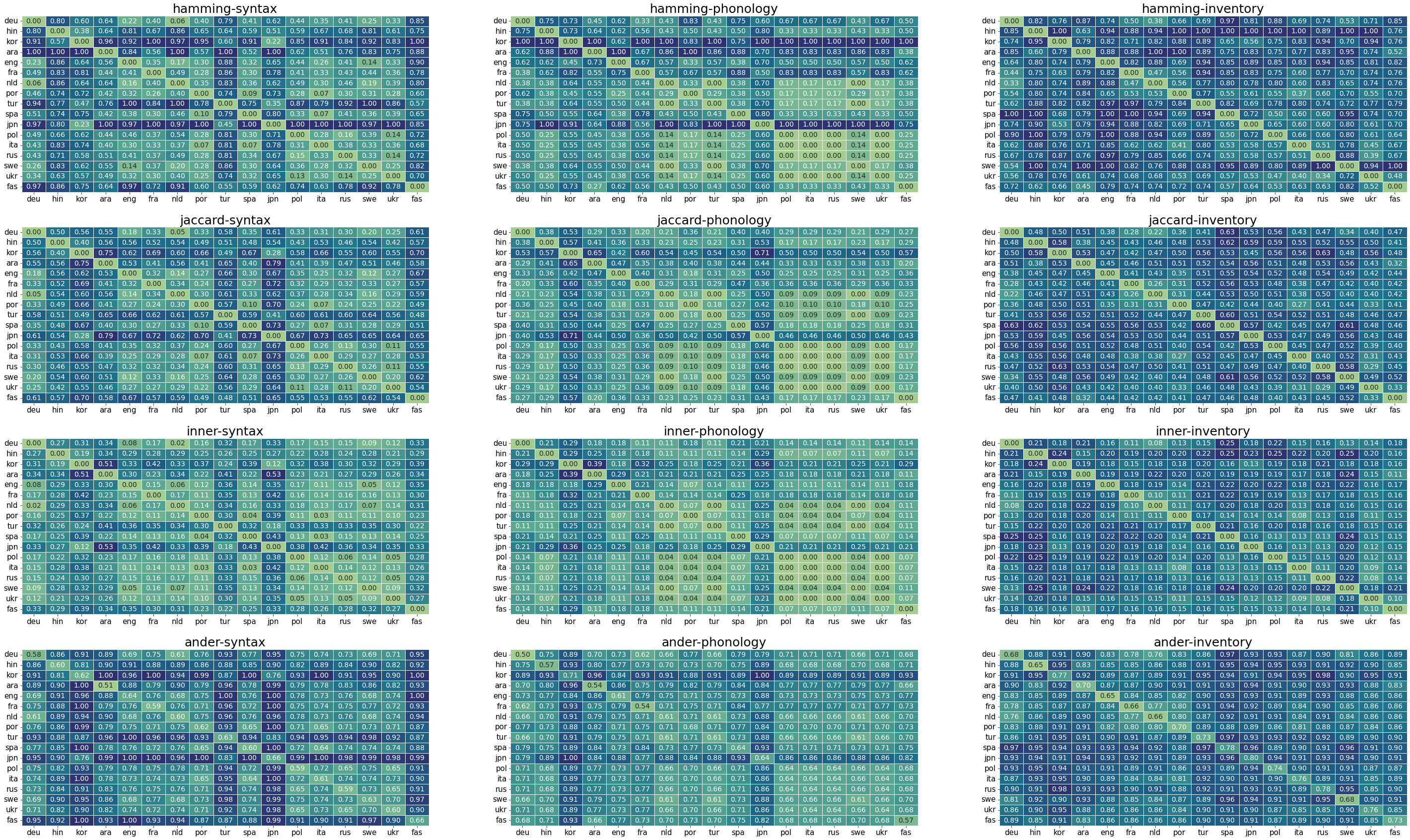}
  \caption{\small Detailed linguistic distances}
  \label{fig:app_lingdist}
\end{sidewaysfigure*}

\begin{figure*}
  \centering
  \includegraphics[width=0.55\textwidth]{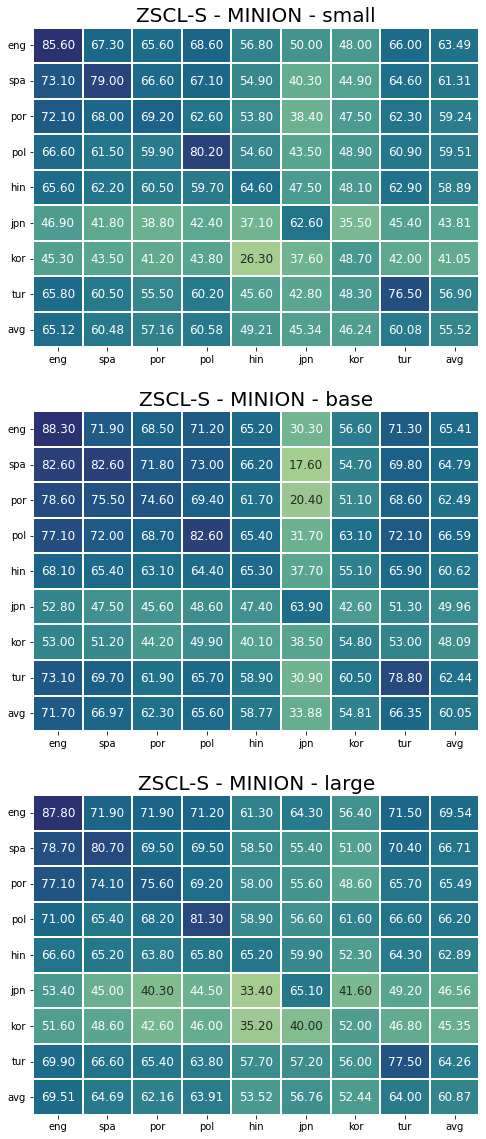}
  \caption{\small Detailed transfer performances for MINION task in ZSCL-S setting.}
  \label{fig:app_zscls_m}
\end{figure*}

\begin{figure*}
  \centering
  \includegraphics[width=0.75\textwidth]{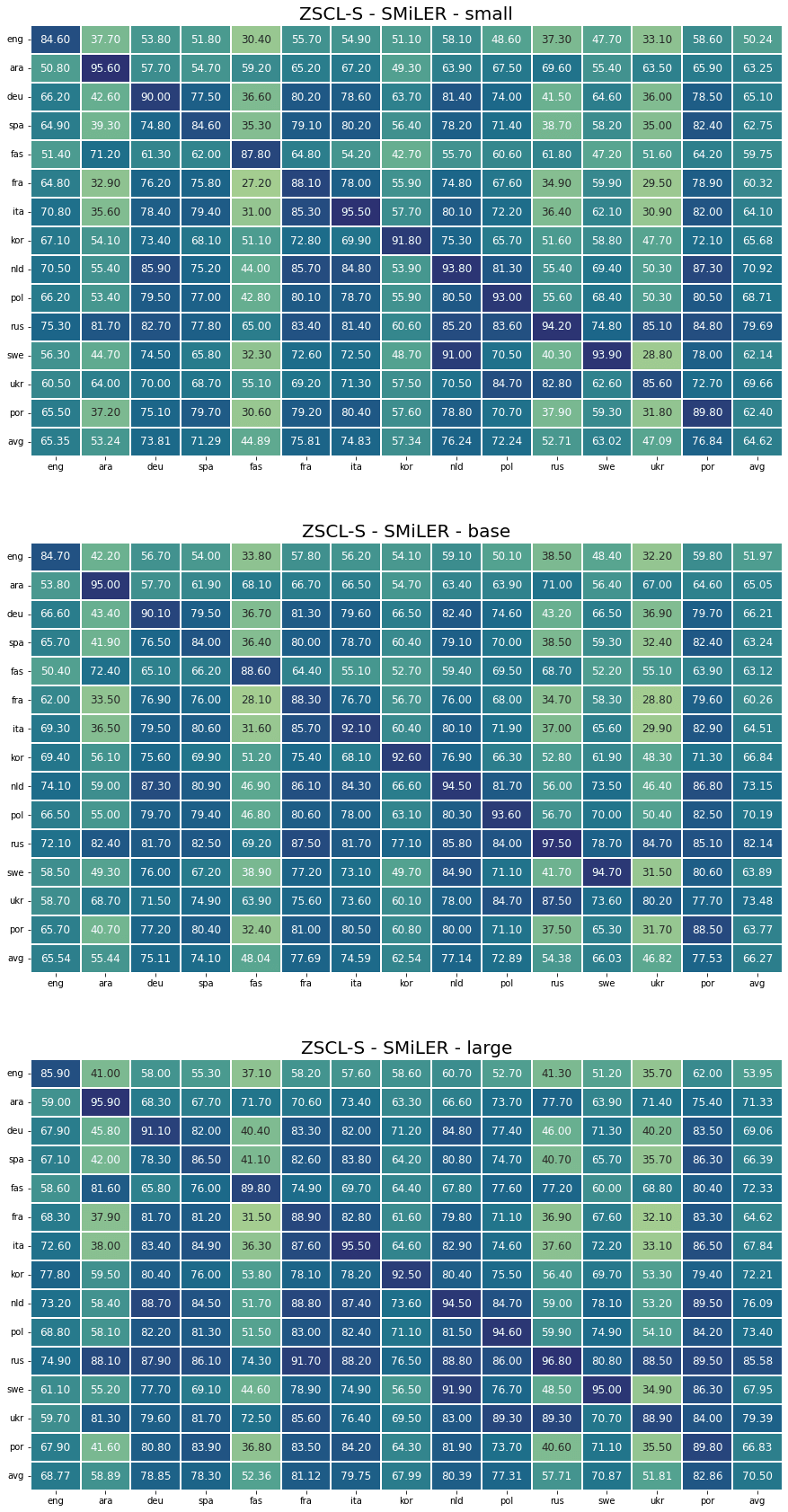}
  \caption{\small Detailed transfer performances for SMiLER task in ZSCL-S setting.}
  \label{fig:app_zscls_s}
\end{figure*}

\begin{figure*}
  \centering
  \includegraphics[width=0.7\textwidth]{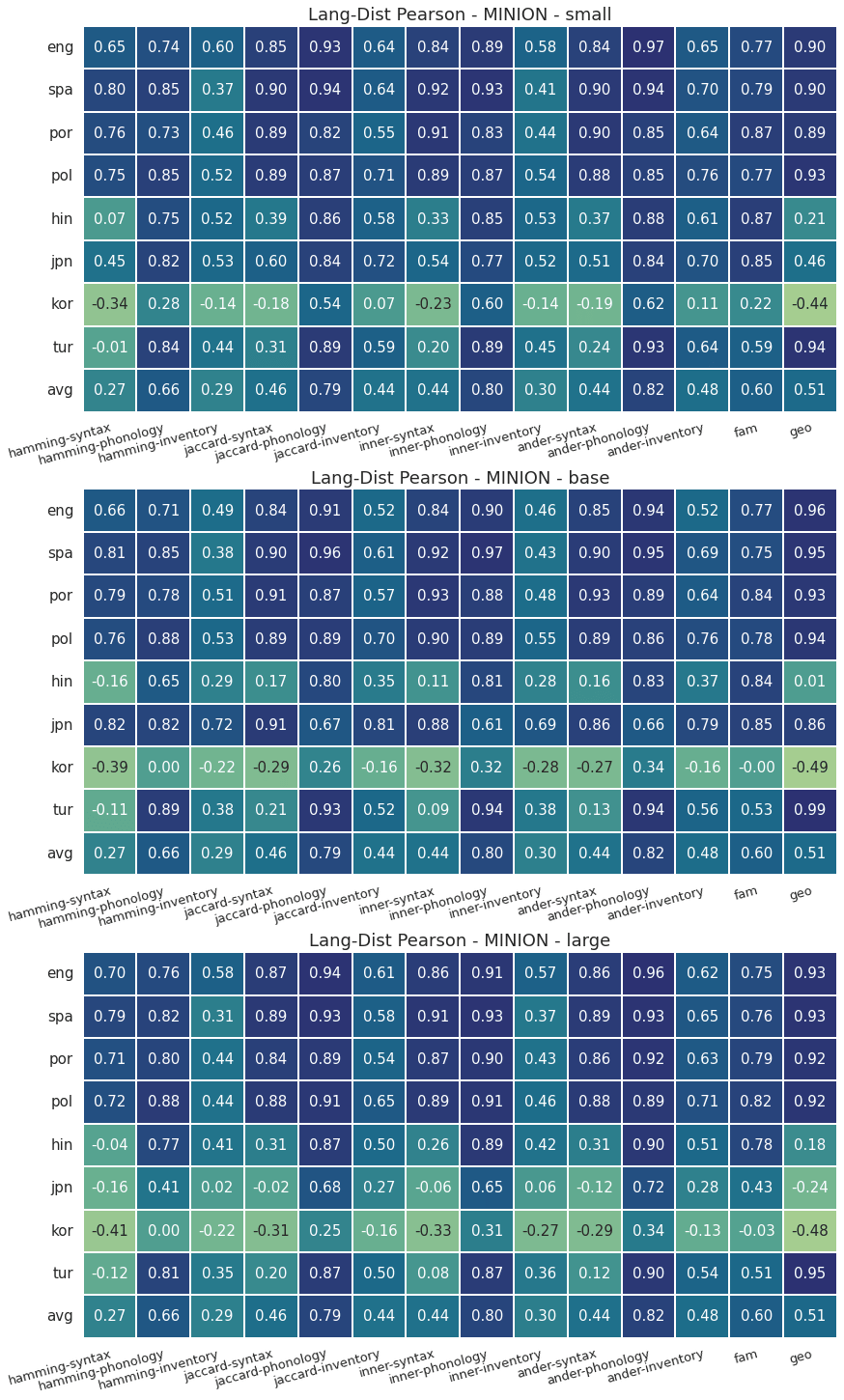}
  \caption{\small Detailed Pearson correlations between ZSCL-S transfer scores and linguistic distances for MINION task.}
  \label{fig:app_corr_zscls_m}
\end{figure*}

\begin{figure*}
  \centering
  \includegraphics[width=0.75\textwidth]{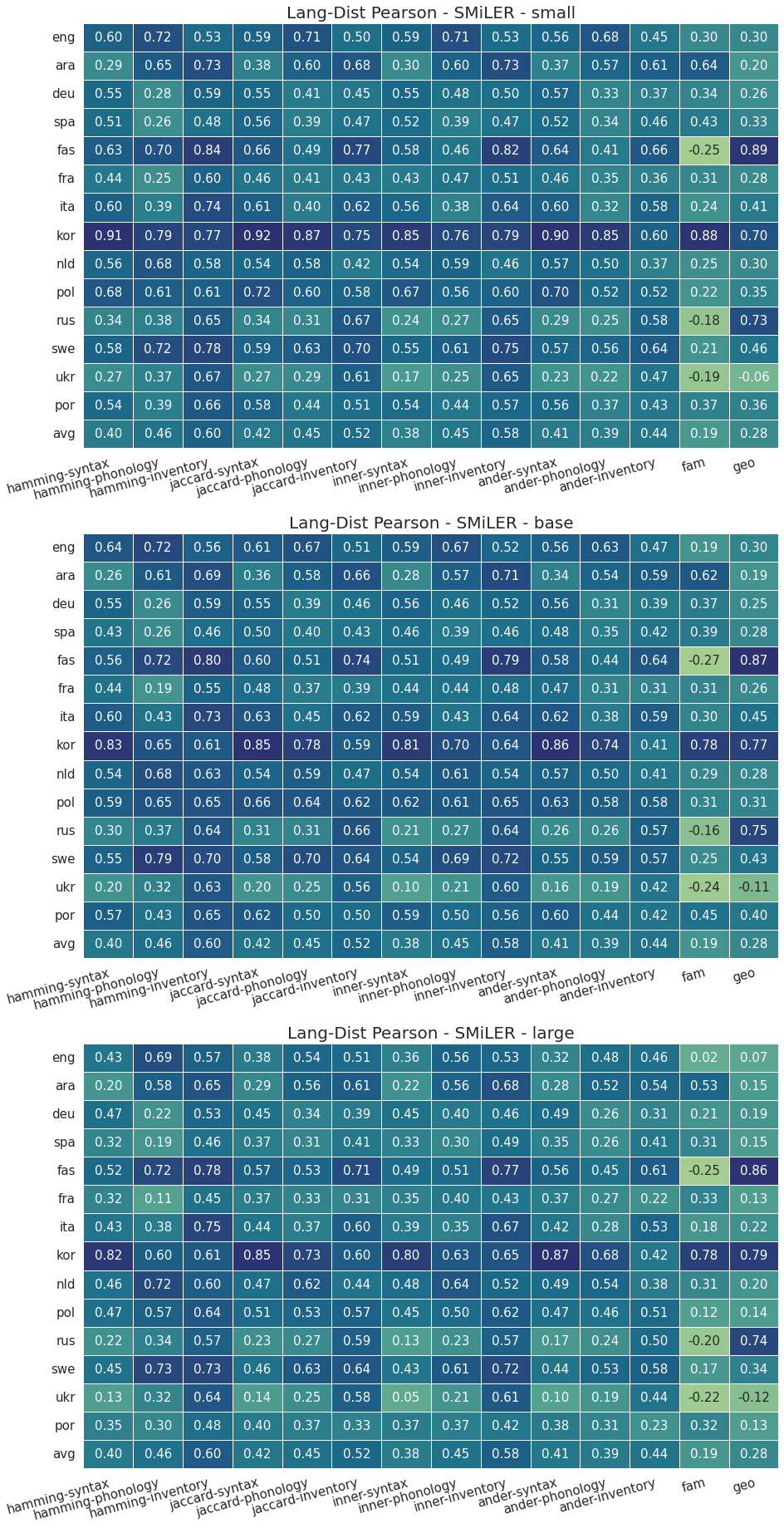}
  \caption{\small Detailed Pearson correlations between ZSCL-S transfer scores and linguistic distances for SMiLER task.}
  \label{fig:app_corr_zscls_s}
\end{figure*}

\begin{figure*}
  \centering
  \includegraphics[width=0.9\textwidth]{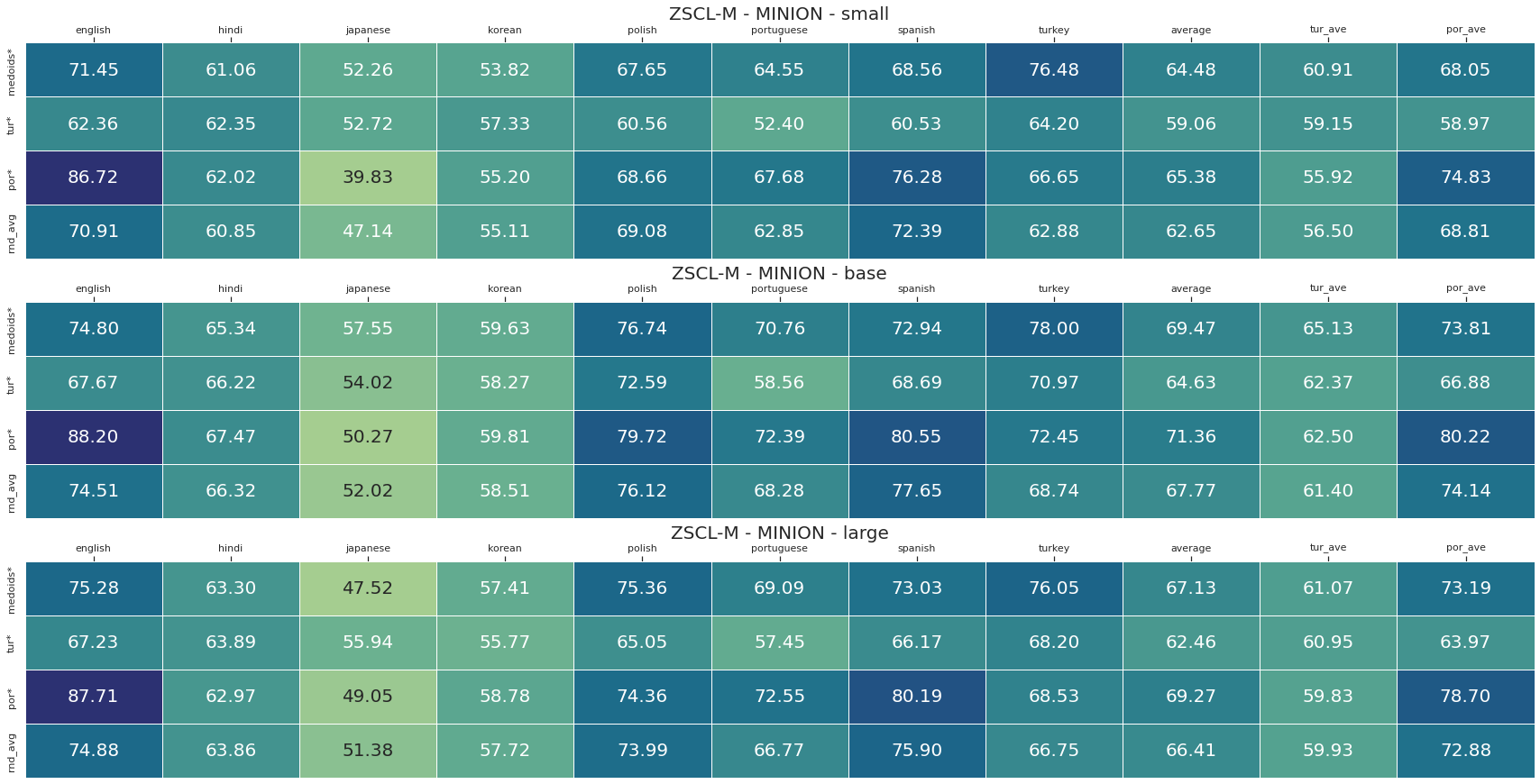}
  \caption{\small Detailed transfer performances for MINION task in ZSCL-M setting.}
  \label{fig:app_zsclm_m}
\end{figure*}

\begin{figure*}
  \centering
  \includegraphics[width=0.95\textwidth]{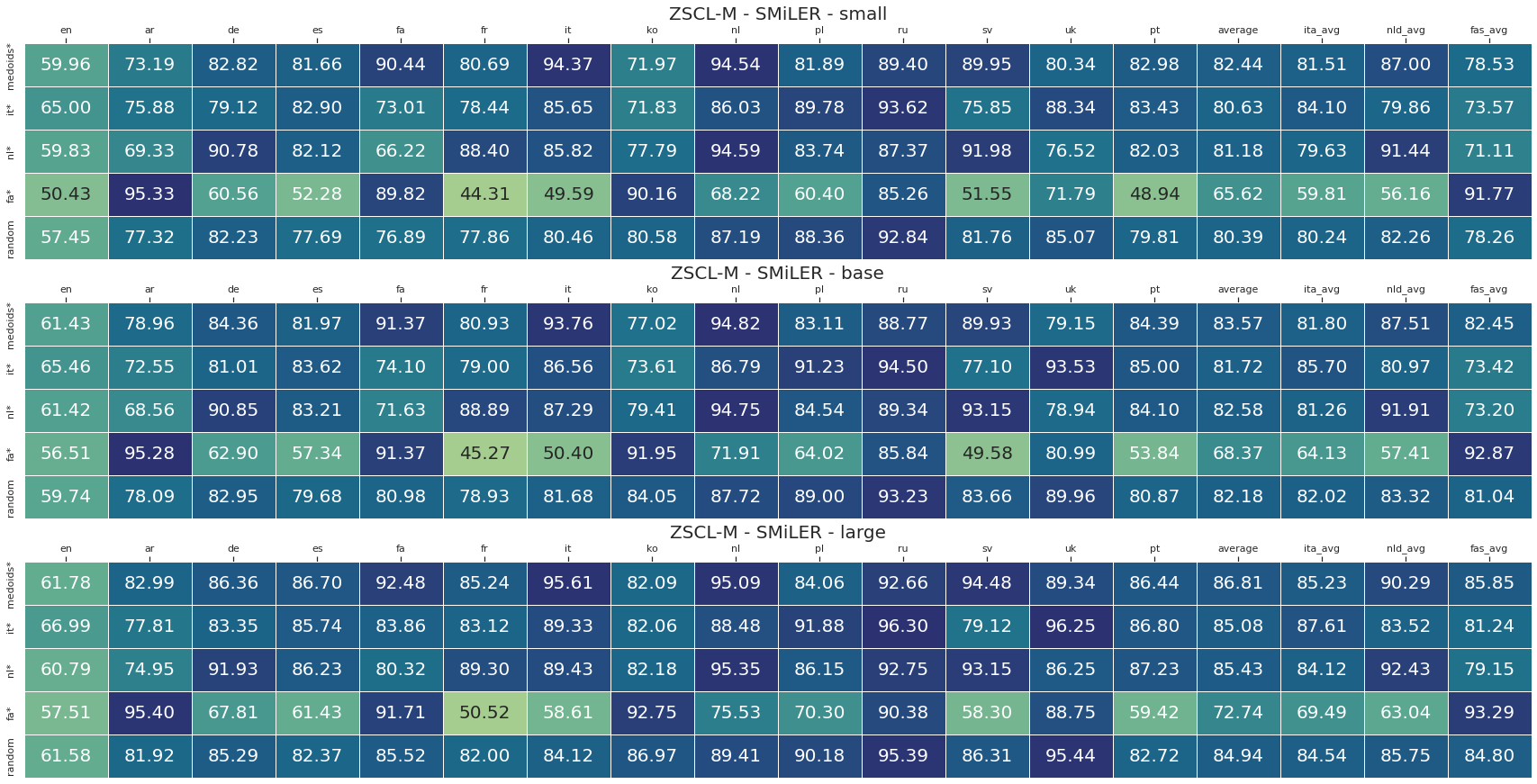}
  \caption{\small Detailed transfer performances for SMiLER task in ZSCL-M setting.}
  \label{fig:app_zsclm_s}
\end{figure*}

\begin{figure*}
  \centering
  \includegraphics[width=0.9\textwidth]{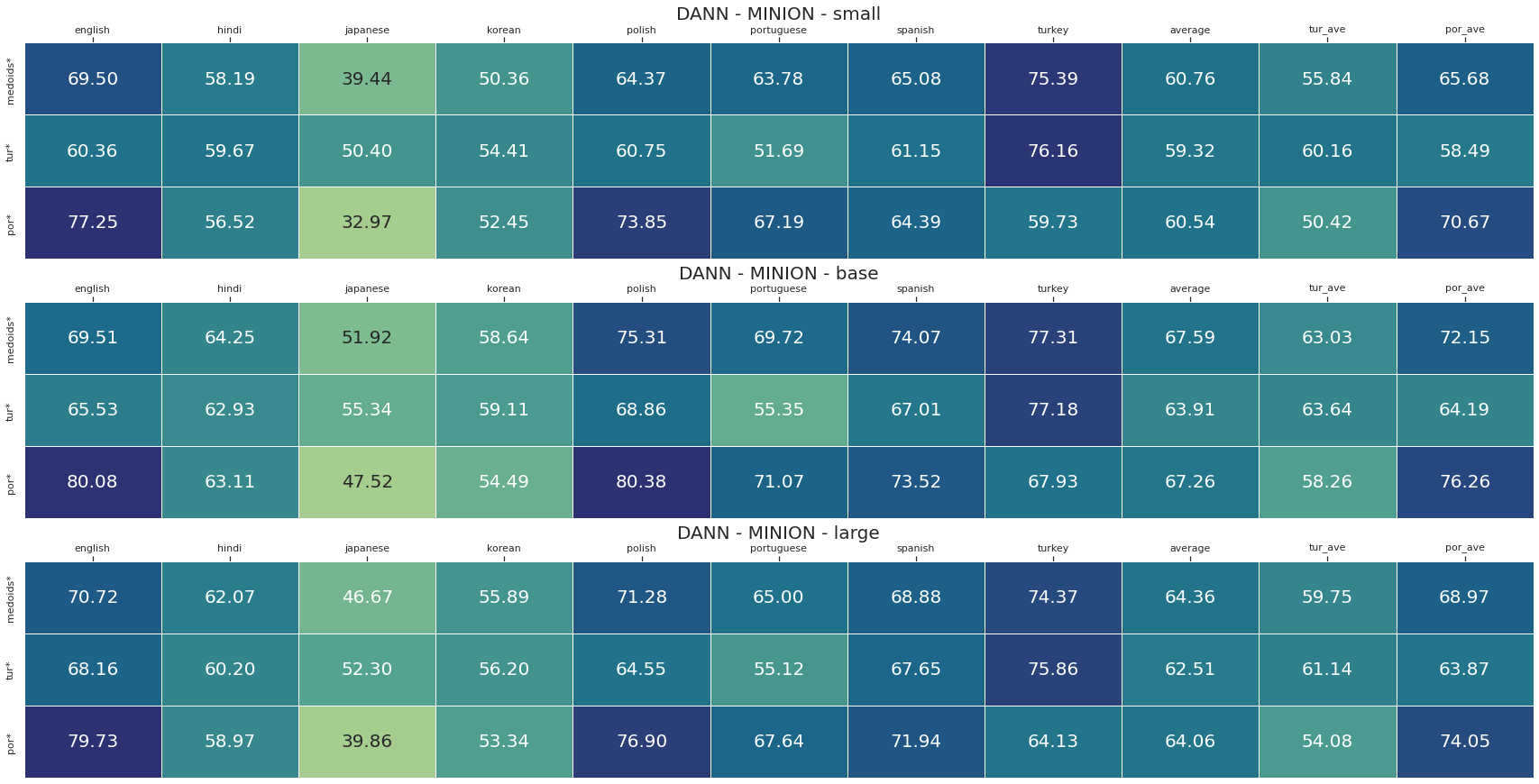}
  \caption{\small Detailed transfer performances for MINION task of DANN baseline.}
  \label{fig:app_dann_m}
\end{figure*}

\begin{figure*}
  \centering
  \includegraphics[width=0.95\textwidth]{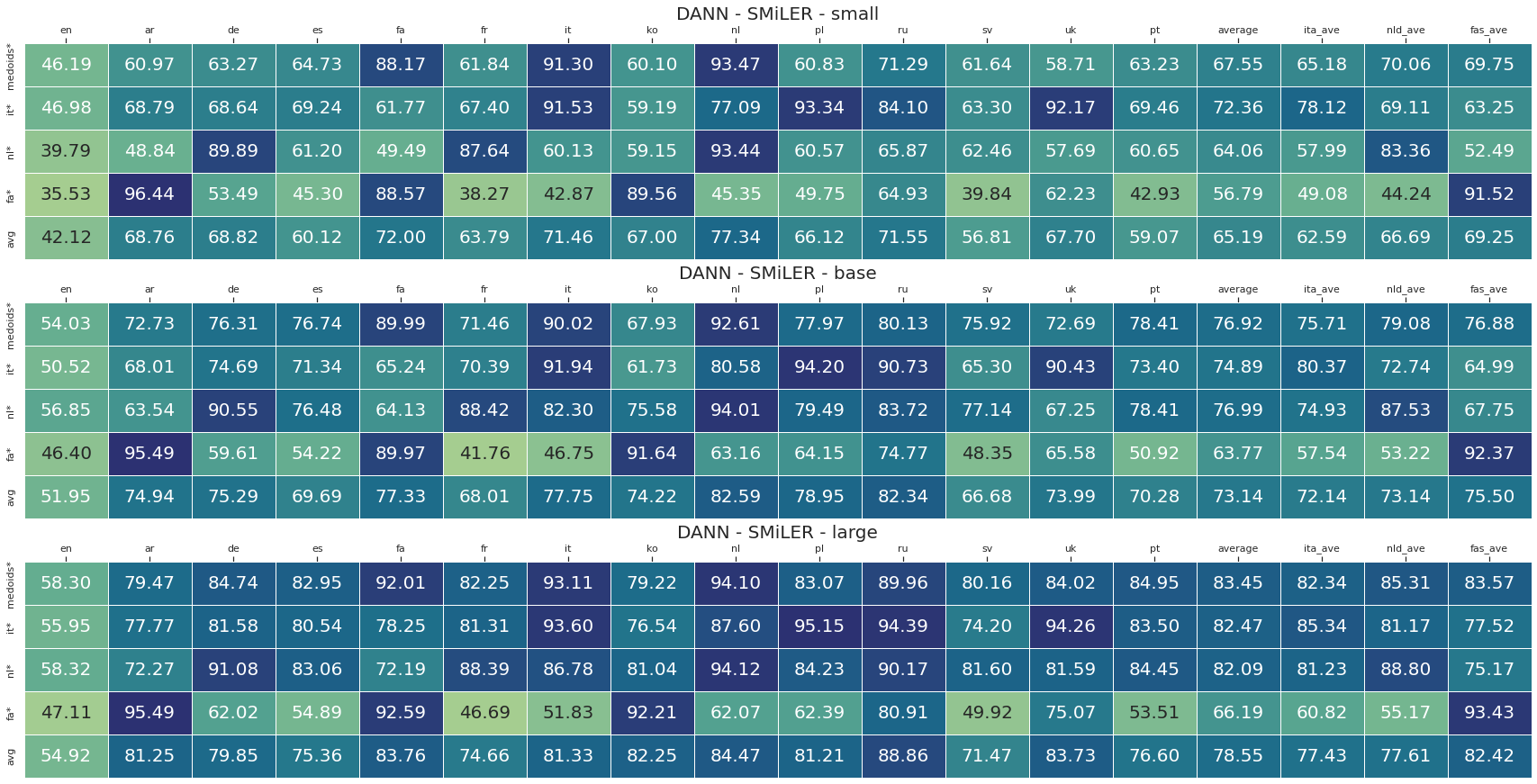}
  \caption{\small Detailed transfer performances for SMiLER task of DANN baseline.}
  \label{fig:app_dann_s}
\end{figure*}

\begin{figure*}
  \centering
  \includegraphics[width=0.9\textwidth]{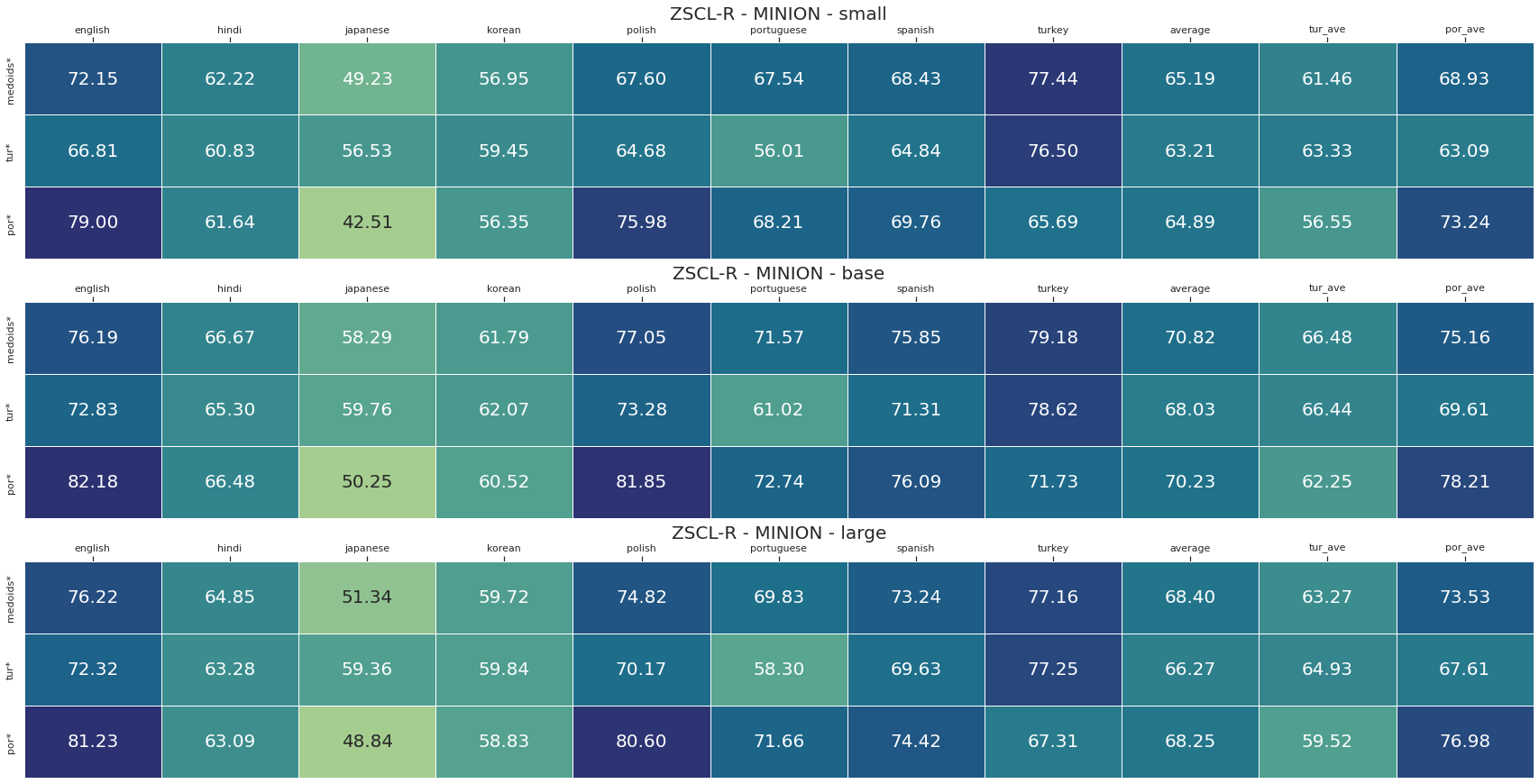}
  \caption{\small Detailed transfer performances for MINION task in ZSCL-R setting.}
  \label{fig:app_zsclr_m}
\end{figure*}

\begin{figure*}
  \centering
  \includegraphics[width=0.95\textwidth]{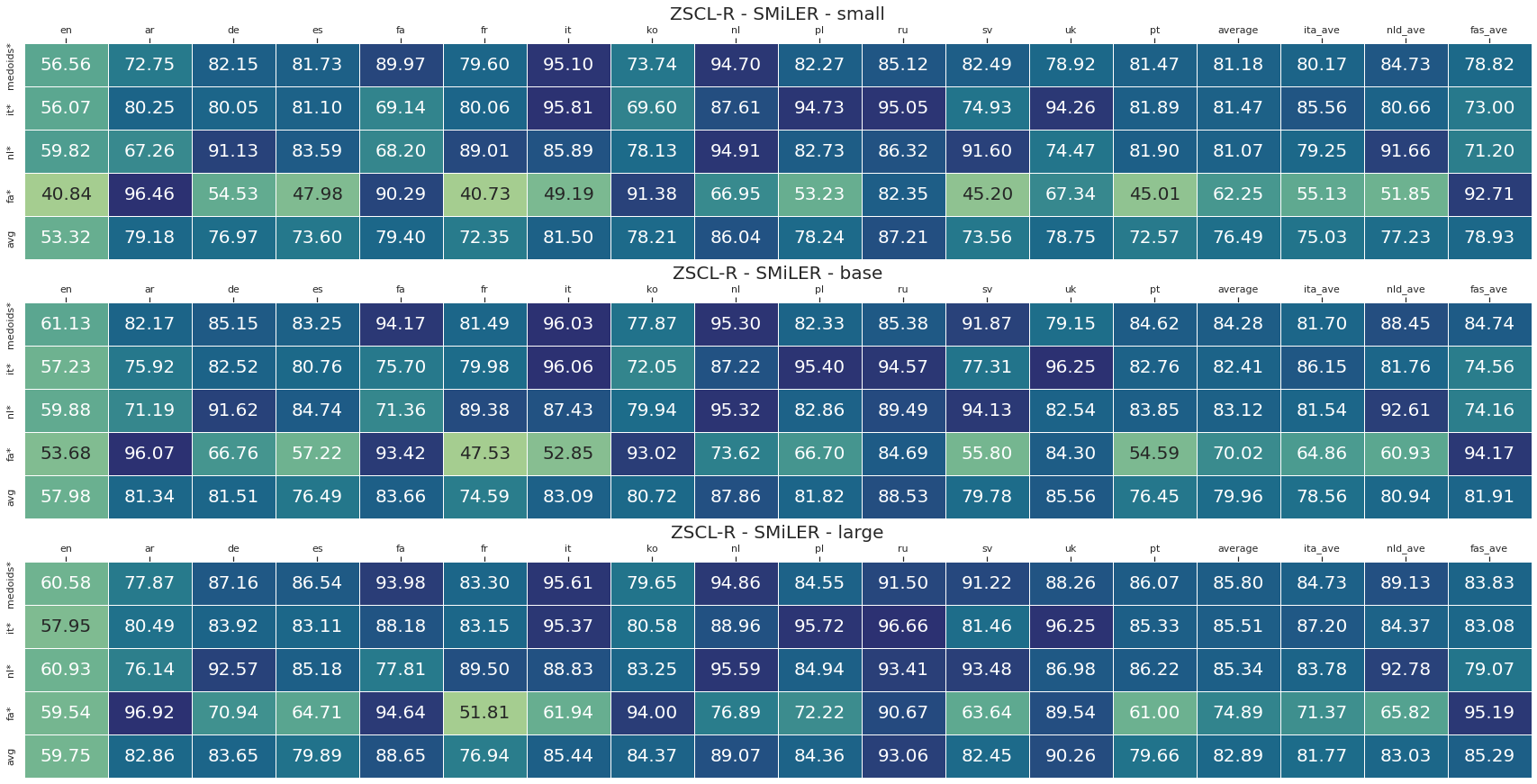}
  \caption{\small Detailed transfer performances for SMiLER task in ZSCL-R setting.}
  \label{fig:app_zsclr_s}
\end{figure*}

In this section, we provide detailed result values for our experiments, including: 
linguistic distances in figure \ref{fig:app_lingdist},
ZSCL-S scores in figures \ref{fig:app_zscls_m} and \ref{fig:app_zscls_s},
transfer-distance correlations in figures \ref{fig:app_corr_zscls_m} and \ref{fig:app_corr_zscls_s},
ZSCL-M scores in figures \ref{fig:app_zsclm_m} and \ref{fig:app_zsclm_s},
DANN scores in figures \ref{fig:app_dann_m} and \ref{fig:app_dann_s},
and finally ZSCL-R scores in figures \ref{fig:app_zsclr_m} and \ref{fig:app_zsclr_s}.

\end{document}